\documentclass{article}

\usepackage{arxiv}

\usepackage{multirow}
\usepackage{float}
\usepackage{natbib}
\usepackage[utf8]{inputenc} 
\usepackage[T1]{fontenc}    
\usepackage{hyperref}       
\usepackage{url}            
\usepackage{booktabs}       
\usepackage{amsfonts}       
\usepackage{nicefrac}       
\usepackage{microtype}      
\usepackage{lipsum}
\usepackage{graphicx}
\usepackage{amsmath}
\usepackage{textcomp}
\graphicspath{ {./images/} }

\title{Investigating Disability Representations in Text-to-Image Models}

\author{
 Yang Tian \\
  Department of Computational Linguistics\\
  University of Zurich\\
  \texttt{yang.tian@uzh.ch} \\
  \And
 Yu Fan \\
  Center for Law \& Economics\\
  ETH Zurich\\
  \texttt{yufan@ethz.ch} \\
   \And
 Liudmila Zavolokina\\
  Department of Information Systems\\
  University of Lausanne\\
  \texttt{liudmila.zavolokina@unil.ch} \\
  \And
 Sarah Ebling \\
  Department of Computational Linguistics\\
  University of Zurich\\
  \texttt{ebling@cl.uzh.ch} \\
}

\begin{document}
\maketitle

\begin{abstract}
Text-to-image generative models have made remarkable progress in producing high-quality visual content from textual descriptions, yet concerns remain about how they represent social groups. While characteristics like gender and race have received increasing attention, disability representations remain underexplored. This study investigates how people with disabilities are represented in AI-generated images by analyzing outputs from Stable Diffusion XL and DALL·E 3 using a structured prompt design. We analyze disability representations by comparing image similarities between generic disability prompts and prompts referring to specific disability categories. Moreover, we evaluate how mitigation strategies influence disability portrayals, with a focus on assessing affective framing through sentiment polarity analysis, combining both automatic and human evaluation. Our findings reveal persistent representational imbalances and highlight the need for continuous evaluation and refinement of generative models to foster more diverse and inclusive portrayals of disability.
\end{abstract}


\section{Introduction}
Techniques for enhancing text-to-image (T2I) foundation models, such as improving aesthetic diversity and image quality, have made astounding progress in recent years \cite{zhou2023vision,li2023multimodal}. State-of-the-art generative models, such as Stable Diffusion \cite{rombach2022high}, DALL·E 3 \cite{dalle33}, GPT-4o Image \cite{gpt-4o-image} and Imagen \cite{saharia2022photorealistic}, achieve remarkable success in producing photorealistic or artistic images from textual descriptions. This success can be attributed largely to training on massive datasets \cite{naik2023social}. For instance, Stable Diffusion is trained on LAION 400M \cite{DBLP:journals/corr/abs-2111-02114} and 5B \cite{schuhmann2022laion}. However, because such training data, comprising both images and texts, are primarily scrapped from the Internet, they inevitably contain problematic content such as stereotypes, offensive or toxic language, and underrepresentation of marginalized groups, leading models to inherit and reproduce various biases \cite{birhane2021multimodal,paullada2021data,mandal2023multimodal}. 

T2I generative models have been shown to exhibit implicit social biases related to gender, ethnicity, skin tone, and geography, such as associating computer scientists with men or defaulting to western settings in everyday scenes \cite{cho2023dall,bianchi2023easily,naik2023social}. However, representations of people with disabilities (PwD) remain largely unexamined, despite their importance in discussions of fairness and ethical AI \cite{trewin2018ai,whittaker2019disability}. 

Beyond the domain of generative AI, disability has been extensively studied in psychology and the social sciences. For example, the stereotype content model \cite{fiske2018model, rohmer2018implicit} suggests that PwD are generally perceived as high in warmth and low in competence \cite{wu2019disability, nario2010cultural}, a view associated with the need for protection and lower social status. As a result, PwD are often stereotyped as vulnerable or passive in social contexts. Much of the research exploring this stereotype has focused on mobility impairments, especially wheelchair users (e.g., \cite{rohmer2018implicit}), thereby implicitly assuming that the ``warm but incompetent'' stereotype applies universally across disability categories. However, recent findings challenge this assumption, showing that different disabilities elicit distinct perceptions of warmth and competence \cite{sadler2015competence, granjon2024disability}. For instance, persons with conditions with psychotic features are often perceived as hostile and incompetent, whereas those with neuro-cognitive conditions tend to be viewed as warm but incompetent \cite{sadler2012stereotypes}. 

Addressing disability representations in T2I models is essential, as inaccurate and stereotypical portrayals can reinforce harmful perceptions and contribute to exclusion. Studying these representations therefore provides an opportunity to advance fairness and inclusivity in AI, while informing broader discussions on the social implications of generative models. Moreover, the extent to which models incorporate representational mitigation techniques offers crucial insight. For example, developers, of DALL·E 3 and GPT-4o Image employ a stack of techniques to avoid risks \cite{dalle3, gpt-4o-image}, whereas models such as Stable Diffusion exhibit relatively loose control on data filtering \cite{sdmodelcard, sdv2}. 

The study at hand investigates both the ways disability is portrayed in T2I models and how discrepancies in the level of representational mitigations influence the portrayal of disability. In particular, we address the following research questions:
\vspace{1em}

\noindent\textbf{RQ1:} How is disability represented in T2I models?

\vspace{0.5em}

\noindent\textbf{RQ2:} How do different representational mitigation strategies influence the depiction of disability in T2I models?

\vspace{1em}

We examine disability representations in T2I models through a two-part experimental design. In Experiment 1, we use a prompt design that incorporates both generic and specific formulations. The generic prompt provides a neutral reference to disability, i.e., ``photo of a person with a disability'', whereas the specific prompts designate distinct disability categories, namely ``photo of a person with mobility impairment'', ``photo of a blind person'', and ``photo of a deaf person''. By calculating image similarity scores, we investigate whether the depiction of a person with a disability is skewed toward a particular category. For this experiment, we select two state-of-the-art T2I models, namely Stable Diffusion XL (SDXL) and DALL·E 3. 

In Experiment 2, we shift the focus from prompt variation to model comparison. We analyze SDXL and DALL·E 3 models, which adopt distinct representational mitigation techniques, to explore how interventions by model developers impact disability representations in T2I models. To analyze how each model represents disability, we combine automatic and human evaluation of sentiment, i.e., whether the generated images convey a more positive, negative, or neutral affective framing. For this analysis, we focus on mental disorders. In addition, we assess how consistently both models portray contrasts between mental disorders and physical/sensory disabilities, examining whether either model tends to depict mental disorders in a more negative framing than physical/sensory disabilities.

The results show that generic representations of disability exhibit the highest similarity with representations of mobility impairment, with SDXL displaying a stronger skew than DALL·E 3. In terms of sentiment, automatic analysis suggested that SDXL produced more negative portrayals, whereas human evaluators judged DALL·E 3’s images as conveying more negativity. A qualitative inspection of the sample images further revealed that DALL·E 3 generated richer and more contextually detailed scenes. While specific contextual elements can enhance emotional depth, they may also reinforce stereotypes associated with certain disabilities.

The remainder of this paper is structured as follows: Section 2 reviews related work, and section 3 outlines the methodology and experiments used to examine disability representations and the effects of representational mitigation techniques. Sections 4 and 5 present the experimental results, discussion, and limitations. Finally, Section 6 concludes with recommendations for improving disability representations in AI-generated content.

\section{Related Work}

\subsection{Disability and Technoableism}
Disability studies scholar Shew \cite{shew2023against} introduced the concept of technoableism to describe ableist assumptions embedded in technological imaginaries. The term reflects a belief in the power of technology that treats the elimination or correction of disability as inherently desirable, reinforcing bias in favor of ways of life of people without disability. Under technoableist framings, technologies designed for disability are often positioned not as optional tools that support diverse ways of living, but as mechanisms of normalization, cure, or correction. 

Shew further argues that AI systems frequently reproduce technoableist assumptions \cite{shew2020ableism}, where AI technologies are often framed as solutions to individual limitations rather than as opportunities to address social or infrastructural barriers. Such framings consider disability as a problem to be managed and PwD as recipients of technological intervention, rather than as agents with diverse desires and lived experiences. For generative models, which synthesize outputs from large cultural data, this perspective highlights how AI can encode and reproduce technoableist imaginaries of disability, which shapes not only who is represented, but how lives of PwD are visually portrayed.

\subsection{Social Representations and Disability in AI Models}

With the rapid growth of research on natural language processing (NLP) systems and their applications, systematic disparities related to gender, race, age, and other social categories remain a significant challenge \cite{blodgett2020language, stanczak2021survey, czarnowska2021quantifying, field2021survey, gupta2023survey, o2024gender, bartl2024gender, fan2025medium, fan2025lexam}. These disparities can emerge at multiple stages, from data collection to model outputs \cite{hovy2021five}, and are commonly addressed through mitigation strategies such as pre-processing inputs and refining model outputs \cite{gallegos2024bias}. A variety of evaluation methods have been proposed and tested in NLP tasks. For example, Czarnowska et al. \cite{czarnowska2021quantifying} examined three transformer-based models to measure representational imbalances in sentiment analysis and named entity recognition. \citep{liang2021towards} developed A-LNLP, a debiasing method based on iterative nullspace projection that does not rely on predefined word lists. \citep{raza2024nbias} introduced ``Nbias'', a framework that incorporates diverse datasets from multiple domains to detect systematic skews. Moreover, interdisciplinary approaches drawing on psychology, behavioral economics, and psychometrics \cite{gupta2023survey, van2024undesirable} also contributed to this discussion.

Building on this broader literature, recent work has begun to examine how PwD are represented in the realm of NLP. For example, researchers have shown that language models often classify sentences containing disability-related terms as negative or toxic \cite{hutchinson2020social, venkit2023automated, li-etal-2024-decoding, wu-ebling-2024-investigating}. An analysis of intersectional ableist bias in the BERT language model further demonstrates that negative associations with disability frequently persist when combined with gender and race identities \cite{hassan2021unpacking}. Research on chatbots also documents disability-related harms \cite{gadiraju2023wouldn, 10.1145/3630106.3658933}. In addition, scholars have proposed various risk assessments regarding other AI systems and their impact on PwD. For example, \citep{guo2020toward} highlighted that in computer vision (CV), facial and body recognition systems may struggle to accurately identify individuals with different facial features or postural differences, potentially leading to discrimination. Similarly, in speech systems, automatic speech recognition may malfunction when used by individuals with atypical speech patterns.

\subsection{Social Representations in Text-to-Image Models}

T2I synthesis has greatly benefited from the CLIP model \cite{radford2021learning,cetinic2022myth}, which enables zero-shot learning by aligning image and text representations. Utilizing CLIP embeddings, DALL·E 2 \cite{ramesh2022hierarchical} has been successful in generating high-quality images using a two-stage model: a CLIP text embedding is passed to an autoregressive or diffusion prior that produces an image embedding, which is then used by a diffusion decoder to generate the final image. DALL·E 3, integrated with ChatGPT, further improves image generation by refining user prompts \cite{dalle33}. Stable Diffusion \cite{rombach2022high} and its successors \cite{sdv2,podell2023sdxl} apply a diffusion-denoising mechanism in the latent space, yielding highly realistic images. Unlike DALL·E, Imagen \cite{saharia2022photorealistic}, Pati \cite{yu2022scaling}, or Midjourney \cite{midjourney}, Stable Diffusion is fully open-source, enabling more comprehensive analyses within the research community. 

Concerns about the social implications of T2I models have grown with their increasing prominence. Studies have shown that social bias can be ingrained in multimodal datasets, thereby interacting with vision and language in the multimodal setting \cite{bhargava2019exposing,ross2020measuring,DBLP:journals/corr/abs-2104-08666,birhane2021multimodal}. Moreover, scaling the size of vision-language datasets has been linked to an increase in harmful and stereotypical content \cite{birhane2023hate}. Several works have examined how T2I models depict social groups along dimensions such as race, gender, age, and geography, using image-text multimodal datasets \cite{ bansal2022well,struppek2023exploiting,naik2023social,bianchi2023easily,cho2023dall}, and media outlets have conducted experiments to highlight similar issues \cite{bloomberg,restofworld}. 

To study these representational patterns, researchers have developed various tools and frameworks. For example, DALL-EVAL analyzes gender and skin-tone distributions by analyzing their distribution in generated images \cite{cho2023dall}. \cite{luccioni2023stable} created textual representations of images and examined the likelihood that a caption or visual question answering (VQA) answer included gender-marked
words. \cite{wang2023t2iat}, inspired by the Implicit Association Test (IAT) \cite{greenwald1998measuring} and the Word Embedding Association Test (WEAT) \cite{caliskan2017semantics}, proposed the Text-to-Image Association Test (T2IAT) framework to detect human biases related to multiple demographic dimensions in image generations. While we do not employ these methods in our study, they exemplify the range of approaches developed to capture systematic patterns in T2I outputs. Another widely used approach is to leverage CLIP-based similarity, which has been shown to surface intersectional associations in language–vision models \cite{wolfe-caliskan-2022-contrastive} and demographic as well as geographical stereotyping in AI-generated images \cite{bianchi2023easily, struppek2023exploiting, urman2025cultural}. Our study draws on this latter line of work, applying similarity measures to disability-related prompts to examine how different disability categories are positioned relative to a generic reference.

\subsection{Disability Representations in Text-to-Image Models}

Compared to NLP and CV, disability in T2I models remains underexplored. A survey-based empirical study offers an early holistic evaluation \cite{mack2024they}. By engaging participants with diverse disabilities to assess a rich collection of generated images depicting disability, e.g., assistive technology (AT), sentiments associated with PwD, the researchers provide a characterization of disability representations. The study highlights negative stereotypes, with PwD often portrayed as wheelchair users, sad, or lonely. In addition, the observations that AT is sometimes more visible than a person with a disability and the lack of diversity in sociodemographic attributes, including gender, age, and race, underscore the need for a broader focus on multiple dimensions of disability representations.

\citep{10.1145/3757887.3763012} advanced this work with a community-driven evaluation, where people with disabilities, health conditions, and accessibility needed co-created prompts to assess T2I outputs. They highlighted how these models often perpetuated harmful stereotypes, defaulted to whiteness, and produced inaccurate or tokenistic depictions of PwD.

Beyond qualitative evaluation, \citep{tevissen2024disability} conducted an experiment on four popular T2I models, statistically analyzing disability representations in their outputs. The findings revealed that PwD were disproportionately depicted as wheelchair users, elderly, and sad, reinforcing stereotypical representations. We expand the scope of analysis by including a wider range of disability categories and by examining both automatic and human evaluations, contributing to a more comprehensive understanding of disability representations.

\subsection{Representational Mitigation Strategies in Text-to-Image Models}
\label{sec:bias-mitigation}

To address representational harms in T2I models, model developers have proposed various strategies. For instance, although DALL·E 2/3 and GPT-4o did not disclose training data, the developers described pre-training mitigations \cite{dalle2pretraining, dalle3}. For the earlier DALL·E 2 model, they first filtered out images with violent and sexual content before training and subsequently re-weighted the filtered dataset after detecting biases induced by the filtering. For example, the training dataset might be more skewed toward presenting women after the filtering process. Thus, the filtered dataset would match the distribution of the unfiltered images after re-weighting. Lastly, the developers employed data deduplication to prevent image regurgitation, a problem where the model reproduces training images verbatim. As an extension, DALL·E 3 and GPT-4o strengthened filtering and added new mitigations. These include blocking measures, such as maintaining textual blocklists, using prompt input classifiers and image output classifiers, and incorporating ChatGPT, which refuses to generate prompts for images with sensitive content. Notably, the developers of DALL·E 3 and GPT-4o Image have explicitly acknowledged applying risk mitigation strategies to diversify portrayals of under-specified groups. 
 
Furthermore, researchers have focused on representational mitigations in T2I models using model weight refinement \cite{wan2024survey}. \citep{struppek2023exploiting} finetuned a text encoder in the model in order to improve model robustness against intentional character-based manipulations in prompts, such as homoglyph substitutions. For instance, replacing the Latin ``o'' in ``photo of an actress'' with a Greek omicron (U+03BF) can cause the model to generate culturally skewed depictions. \citep{orgad2023editing} proposed TIME (Text-to-Image Model Editing), a method that efficiently updates a model’s cross-attention layers to align an under-specified prompt with a desired attribute. A different approach focuses on improving text prompts \cite{prerak2024addressing}. Researchers mitigated skewed representations of gender, age, race, and geographical location that depict occupations, personality traits, and everyday situations by increasing the amount of specificity in the prompts \cite{naik2023social}. Building on this, \citep{esposito2023mitigating} proposed training on synthetic prompts that combine various demographic attributes to mitigate systematic skews.

While work on social representations in NLP and T2I models has focused mainly on gender and race, disability remains underexplored. Moreover, studies on representational mitigation rarely assess whether such interventions introduce new skews or reinforce stereotypes.

The present study addresses these gaps by conducting an analysis of disability representations in T2I models, focusing on two core aspects. First, we examine whether models default to mobility impairment as the primary representation of disability. Second, we assess how representational mitigation techniques influence these depictions, comparing models with distinct levels of representational mitigations.

\section{Methodology}

Our study consists of two experiments. Experiment 1 examines representations of specific categories of disability in AI-generated images relative to generic representations. Experiment 2 evaluates the effect of representational mitigation techniques, with a particular focus on the sentiment polarity of the images. 

\subsection{Experiment 1: Detection on Disability Representation Differences}

The aim of this experiment is to investigate disability representations in images generated by T2I models by quantifying their differences. We use a generic prompt, ``photo of a person with a disability'', as a reference to assess how the models visually interpret PwD. This neutral phrasing is deliberate: it allows the default assumption of the model to surface by leaving context unspecified. In other words, it reveals which disability categories it depicts most often without external control. We also did not add any descriptors, such as occupations, in the prompt, as they would direct the model toward a specific context, potentially masking underlying stereotypes. This approach is consistent with methodologies in prior work that used neutral prompts to uncover imbalanced representations in generative models \cite{naik2023social}.

In addition, we generate images from specific prompts to capture potential deviations from the generic representation. The specific disability categories are chosen to include the following types: mobility impairment, blindness, and deafness. They are selected based on their recognizable visual characteristics, such as assistive devices and service animals, which are likely to be captured by image generation models without requiring explicit specification in the prompt. We intentionally exclude disabilities without clear visual cues, such as chronic illnesses or certain mental health conditions, because their depiction in static images is often based on atmosphere or setting rather than distinct physical markers of the persons in the images. 

\subsubsection{Data}
This study focuses on two T2I models: SDXL and DALL·E 3. All images were created in July 2025. The DALL·E 3-generated images were obtained using the official OpenAI API \footnote{https://platform.openai.com/docs/guides/image-generation?image-generation-model=dall-e-3}. We used the API’s default image generation parameters. Details are listed in Appendix \ref{appendix:dalle3} for reproducibility. For each prompt, both generic and specific, each model generated 100 images, resulting in a dataset of 800 images in total.

\subsubsection{Evaluation} 
\label{sec:exp1-eval}

\begin{figure}[ht]
\vskip 0.1in
\begin{center}
  \includegraphics[width=13cm]{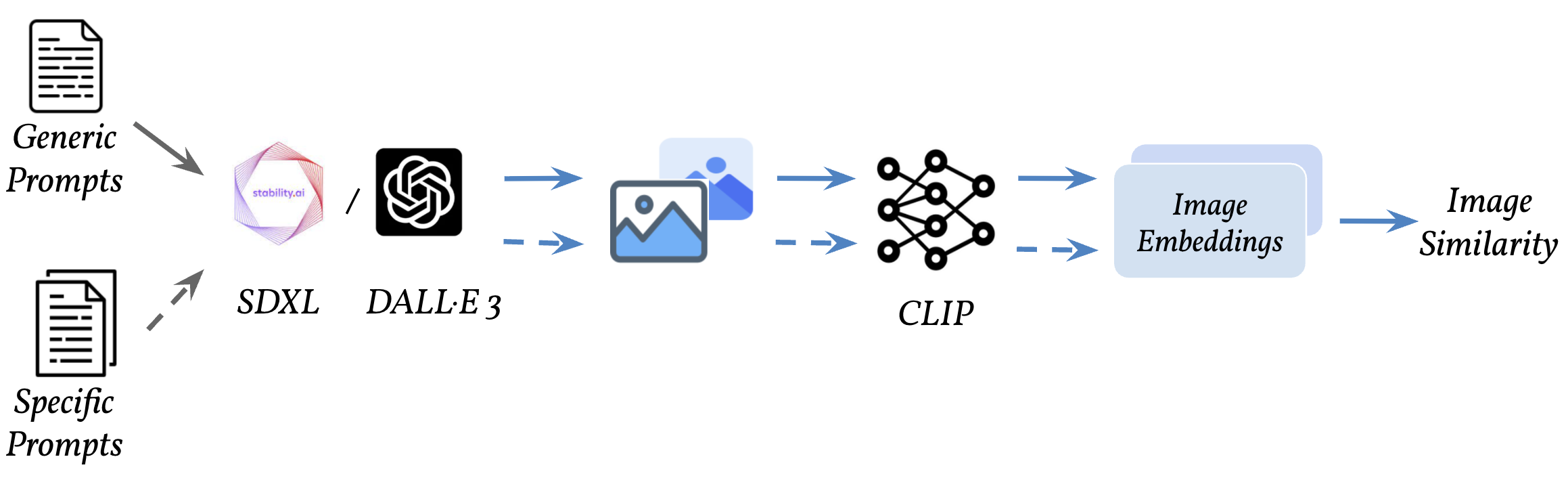}
    \caption{\small Overview of the evaluation process for detecting disability representation differences. Image similarity between generated images using generic and specified prompts is measured with CLIP embeddings.}
  \label{fig:img/bias}
  \end{center}
\vskip -0.1in
\end{figure}

To quantify representational differences, we calculate similarity scores between the generated images. Figure \ref{fig:img/bias} outlines the process. First, we input both generic and specific prompts into the models to generate images. Then, we extract CLIP embeddings \cite{radford2021learning} for each image and compute the cosine similarity between them.

We use CLIP similarity not as a definitive measure of sociocultural bias but as a relative indicator of deviation from a general reference representation, consistent with prior work showing that CLIP-based similarity can capture underlying representational skews \cite{struppek2023exploiting, urman2025cultural}. In this experiment, higher similarity scores indicate that images from a specific disability category are visually closer to the generic prompt, suggesting that this category may be over-represented as the ``default'' depiction of disability. Lower scores indicate greater visual divergence from the generic prompt.

While CLIP embeddings capture high-level semantic content, such as the presence of assistive devices or characteristic postures, they also encode lower-level visual cues like color palettes, lighting, and backgrounds. This means that two images could receive high similarity scores due to superficial style rather than substantive content. To reduce this risk, we focus on disability categories with identifiable visual markers, such as wheelchairs, white canes and hearing aids, while allowing diversity in other aspects of the images. This increases the likelihood that our similarity scores reflect meaningful content related to disability representations rather than stylistic similarities. We therefore interpret these scores as a comparative metric that highlights which categories are positioned closer or further from the generic representation, rather than as an absolute measure of bias.

To quantify how closely the generic prompt aligns with each disability category, we compute, for each generic-prompt image $\mathit{i}$ and category $\mathit{c}$:

\begin{equation}
\Delta_i^{(c)} = \mathrm{sim}_i(\text{generic}, c) - \frac{1}{K-1} \sum_{\substack{c' \neq c}} \mathrm{sim}_i(\text{generic}, c')
\end{equation}

where $\mathrm{sim}_i$ is the cosine similarity between CLIP embeddings of generic-prompt image $\mathit{i}$ 
and the images from category $\mathit{c}$, $\mathit{K}$ is the total number of categories, and $\mathit{c'}$ indexes all categories other than $\mathit{c}$. 
A positive $\Delta_i^{(c)}$ indicates that the generic prompt is, on average, more visually similar to category $\mathit{c}$ than to the other categories; a negative value indicates the opposite.

We then aggregate $\Delta_i^{(c)}$ across all generic-prompt images to compute a mean similarity score for each category and estimated 95\% bootstrap confidence intervals to assess the stability of these category-level effects.

\subsection{Experiment 2: The Effect of Representational Mitigation Techniques on Disability Representations}

The second experiment investigates the influence of mitigations and concentrates on the comparison between SDXL and DALL·E 3, which exhibit significant disparity in representational mitigation techniques. Whereas the Stable Diffusion model is primarily trained on the uncurated large LAION dataset \cite{DBLP:journals/corr/abs-2112-10752} and the training data is solely filtered using a NSFW detector \cite{sdmodelcard}, DALL·E 3 applies extensive pre-training mitigations \cite{dalle3} (see Section~\ref{sec:bias-mitigation}).

To capture the potential impact of these strategies, we consider sentiment polarity of the generated images as our primary indicator of affective framing, i.e., whether an image conveys a predominantly positive, neutral, or negative tone. In addition, we examine specific emotions and broader atmosphere cues, such as facial expressions and background context, since these fine-grained affective elements can shape how sentiment is perceived and provide complementary insights into how disabilities are framed in the outputs.

Mental disorders are a natural focus for this experiment, as they are often associated with shifts in affect, making sentiment analysis particularly relevant for evaluating their representations. We include three categories: bipolar disorder\footnote{Bipolar disorders are ``[...] mental health conditions characterized by periodic, intense emotional states affecting a person's mood, energy, and ability to function'' \cite{bpdefinition}. In the ICD-11, the disease classification system maintained by the World Health Organization (WHO), bipolar disorders are classified under code 6A6, with Bipolar I Disorder (6A60), Bipolar II Disorder (6A61), and Cyclothymic disorder (6A62) specifically identified \cite{ICD-11}. In the DSM-5-TR, the diagnostic classification system published by the American Psychiatric Association, bipolar and related disorders are placed between the chapters on schizophrenia spectrum and other psychotic disorders and depressive disorders, highlighting their role as a bridge between these diagnostic categories in terms of symptomatology, family history, and genetics \cite{american2013diagnostic}.}, depression\footnote{Depression [...] ``is a common and serious mental disorder that negatively affects how you feel, think, act, and perceive the world'' \cite{dedefinition}. In the ICD-11, depressive disorders are classified under code 6A7 \cite{ICD-11-de}. In the DSM-5-TR, depressive disorders are separated from and placed behind bipolar disorders. They differ in duration, timing, or presumed etiology \cite{american2013diagnostic}.}, and anxiety disorders\footnote{Anxiety disorders are the most common mental disorders. They ``[...] differ from normal feelings of nervousness or anxiousness and involve excessive fear or anxiety \cite{addefinition}. In the ICD-11, anxiety or fear-related disorders are classified under code 6B0 \cite{ICD-11-ad}. In the DSM-5-TR, anxiety disorders are placed behind depressive disorders \cite{american2013diagnostic}.}, using a prompt ``photo of a person with bipolar disorder'' and analogous for the other two categories.

In addition, we extend the analysis to compare mental disorders with physical/sensory disabilities. Unlike mental disorders, conditions such as blindness, deafness, and mobility impairment are not inherently linked to an emotional state and thus provide an important point of contrast. This distinction raises the possibility that models may rely more heavily on negative sentiment framing when generating images of mental disorders.

\subsubsection{Data} 
For the mental disorder categories (anxiety disorder, depression, bipolar disorder), we collected 300  images per category from each model (SDXL and DALL·E 3), resulting in a total of 600 images. For the physical/sensory categories (mobility impairment, blindness, deafness), we use the 600 images already collected in Experiment 1 (300 per category per model).

We use two complementary evaluation methods. First, we apply automatic sentiment polarity analysis. Second, we conduct a human evaluation using pairwise comparisons to capture perceived differences in sentiment between models and between disability groups.

For automatic evaluation, we use all 1,200 images across the six disability categories and two models. For human evaluation, we construct two sets of paired comparisons:

\begin{enumerate}
    \item Model comparisons: From the mental disorder group (three categories, 300 images per model), we randomly sample 30 images from each model, pairing each DALL·E 3 image with an SDXL image from the same category. This yields 30 model comparison pairs.
    
    \item Group comparisons: For each model, we randomly sample 15 images from the mental disorder group (out of a total of 300 images) and 15 images from the physical/sensory group (total: 300 images), pairing them by random assignment. This yields 15 group comparison pairs per model and 30 group comparison pairs in total.   
\end{enumerate}

In total, the human evaluation includes 60 unique image pairs, plus three repeated pairs (5\% of total) to assess intra-rater reliability. Because human evaluation is more time-consuming than automatic sentiment analysis, we restrict it to this subset of the full dataset for this task. The subset is designed to be representative of each category and model while keeping the task manageable for evaluators.

\subsubsection{Automatic Evaluation} 

\begin{figure}[ht]
\vskip 0.1in
\begin{center}
  \includegraphics[width=13cm]{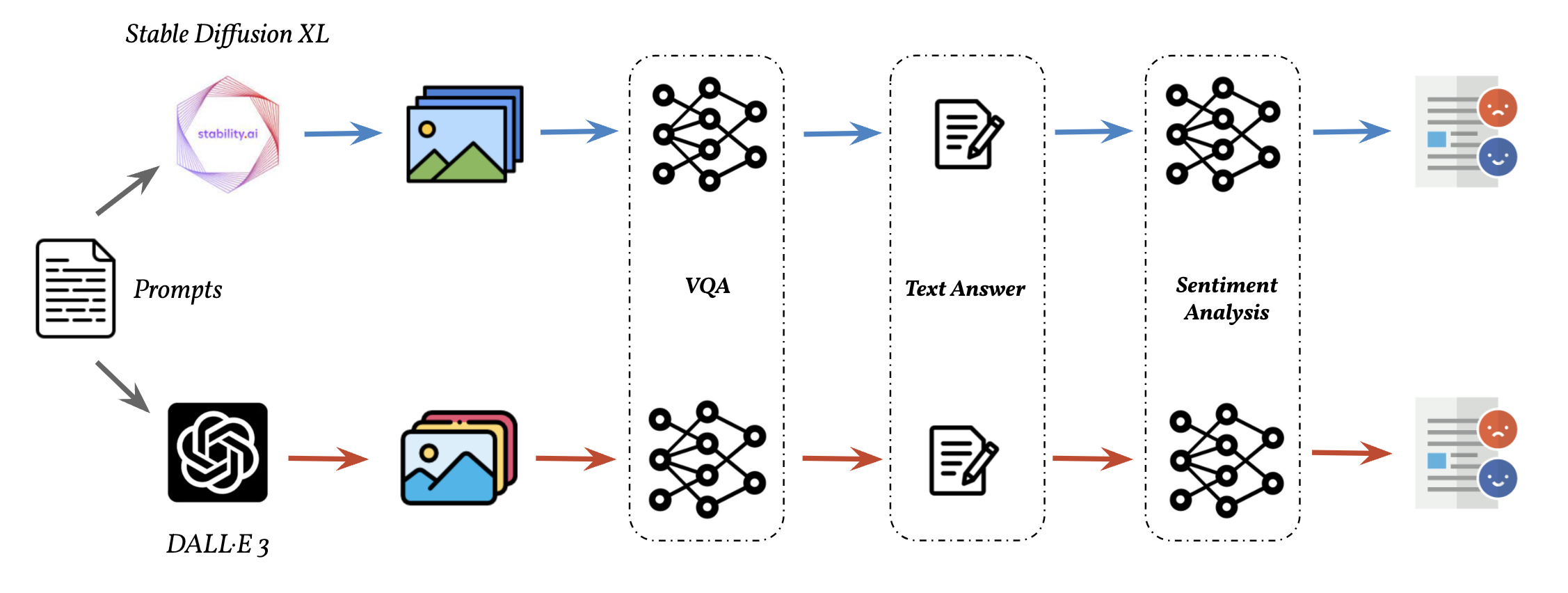}
    \caption{\small Overview of the evaluation process for detecting sentiment of synthesized images from different models. The comparison is made between Stable Diffusion XL and DALL·E 3.}
  \label{fig:flow-disability2}
  \end{center}
\vskip -0.1in
\end{figure}

We adopt a two-step pipeline, as illustrated in Figure \ref{fig:flow-disability2}. First, we extract text-based sentiment descriptions of the images using the BLIP VQA system, which has demonstrated strong performance in producing semantically rich textual descriptions that align well with CLIP's visual–semantic space \cite{li2022blip}, making them suitable for downstream NLP tasks. For each image, we ask:

 \begin{enumerate}
     \item ``How would you describe the emotional atmosphere created by the background and setting of the image?''
     \item ``What overall mood or feeling does the image convey as a whole?''
     \item ``What emotion is visible on the person's face or in their body language?''
 \end{enumerate}

The first two queries capture scene-level emotional framing, while the third targets individual-level expressions. Together, these responses provide the basis for our sentiment polarity analysis. The extracted texts are analyzed using a sentiment classifier to assign polarity labels (positive, neutral, negative), specifically the \textit{twitter-roberta-base-sentiment-latest} model developed by Cardiff NLP \cite{camacho-collados-etal-2022-tweetnlp, loureiro-etal-2022-timelms}. We then aggregate the results to compare sentiment distributions between models and across disability groups, which enables us to identify systematic differences in how each one was portrayed.

We treat automatic sentiment analysis as an exploratory measure that was useful for detecting broad patterns but not definitive on its own. We complement it with human evaluation. This allows us to compare patterns between automatic and human judgments and assess the robustness of our findings.

\subsubsection{Human Evaluation}

We recruit three evaluators using Prolific \cite{prolific} to assess their sentiment judgments in the 63 image pairs. Evaluators were drawn from the general Prolific pool, without demographic restrictions or screening for disability status. Each pair is evaluated by all three evaluators, allowing us to measure inter-rater agreement. The order of all pairs is randomized in the survey. For each pair, evaluators answer the question: ``Please select the image that you feel conveys a more negative emotion or mood, based on the emotions of the person, the atmosphere of the scene, or any other emotional/atmospheric cues you perceive.''

After making a choice, evaluators are also asked to rate their confidence in that choice on a scale from 1 (``Not at all confident'') to 5 (``Very confident''). Combining agreement and confidence provides a richer picture of how evaluators perceive differences between images. For example, high agreement with low confidence may indicate that the visual differences between two images are subtle, and evaluators noticed something similar but are uncertain about its strength. Low agreement with high confidence could suggest that evaluators interpret the same images in fundamentally different ways and are sure of their own perception but focus on different visual or emotional cues.

\section{Results}

\subsection{Experiment 1: Similarity of Generic and Specific Disability Representations}

\begin{figure}[ht]
\vskip 0.1in
\begin{center}
  \includegraphics[width=15cm]{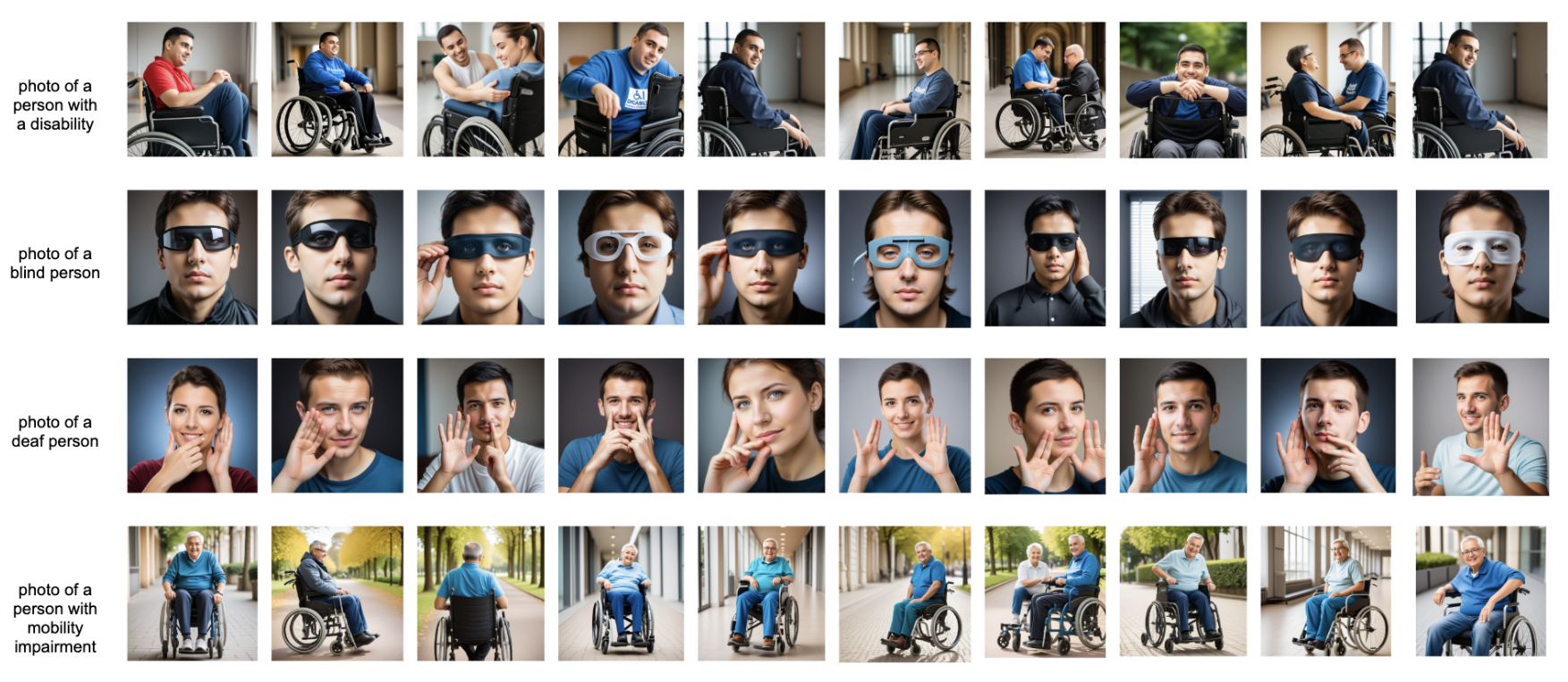}
    \caption{\small Sample images generated from the generic and specified prompts using Stable Diffusion XL}
    \label{fig:sdxl_disability}
  \end{center}
\vskip -0.1in
\end{figure}

\begin{figure}[ht]
\vskip 0.1in
\begin{center}
  \includegraphics[width=15cm]{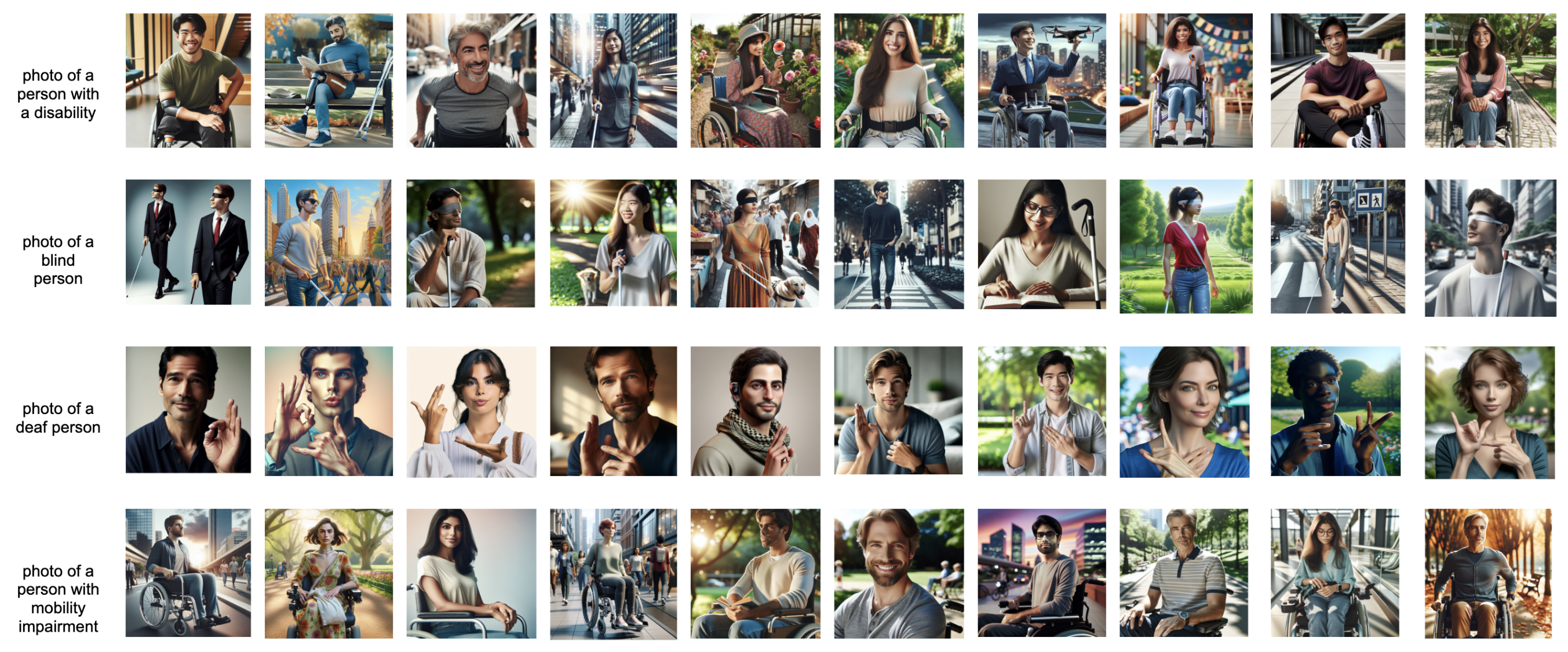}
    \caption{\small Sample images generated from the generic and specified prompts using DALL·E 3}
  \label{fig:dalle3_disability}
  \end{center}
\vskip -0.1in
\end{figure}

The first experiment examined similarity in representations across generic and specific prompts. Figures \ref{fig:sdxl_disability} and \ref{fig:dalle3_disability} show sample images generated by SDXL and DALL·E 3, respectively. The images in the first row were generated using the generic prompt as a reference, while the subsequent rows depict specific disability categories. 

\subsubsection{Descriptive Statistics}

Tables \ref{table:simsd} and \ref{table:simdalle3} provide the similarity scores plus associated statistics based on the CLIP embeddings of images generated using the generic and specific prompts. High similarity scores indicate that images representing specific disabilities are very similar to those generated using the generic prompt. Overall, the gap between the similarity scores, i.e., mean, minimum, and maximum, in each group is small, with low standard errors, likely because both models tended to produce homogeneous human figures or backgrounds across images, as is visible from Figures \ref{fig:sdxl_disability} and \ref{fig:dalle3_disability}. The highest similarity was consistently observed between the generic prompt and mobility impairment, suggesting that when asked to depict ``a person with a disability'', models tended to default to mobility impairment over other categories.

\begin{table}
  \caption{Descriptive statistics of the cosine similarity scores between CLIP embeddings of the synthesized images with generic and specific prompts from Stable Diffusion XL. The number of observations of each group pair is 10,000 (100$\times$100).}
  \label{table:simsd}
  \renewcommand{\arraystretch}{1.25}
  \small
  \begin{tabular}{|p{7cm}|r|r|r|r|r|}
    \hline
    \textbf{Generic \& Specific Prompt} & \textbf{Mean} & \textbf{SD} & \textbf{Min} & \textbf{Max} & \textbf{Range (Max - Min)} \\
    \hline
    \textit{a person with a disability} \& \textit{a person with mobility impairment} &  0.9013 & 0.0302 & 0.8047 & 0.9785 & 0.1738 \\
    \hline
    \textit{a person with a disability} \& \textit{a blind person} & 0.8081 & 0.0288 & 0.7173 & 0.8740 & 0.1567 \\
    \hline
    \textit{a person with a disability} \& \textit{a deaf person}  & 0.8229  & 0.0279 & 0.7241 & 0.9106 & 0.1865 \\
    \hline
  \end{tabular}
\end{table}

\begin{table}
  \caption{Descriptive statistics of the cosine similarity scores between CLIP embeddings of the synthesized images with general and specific prompts from DALL·E 3. The number of observations of each group pair is 10,000 (100$\times$100).}
  \label{table:simdalle3}
  \renewcommand{\arraystretch}{1.25}
  \small
  \begin{tabular}{|p{7cm}|r|r|r|r|r|}
    \hline
    \textbf{Generic \& Specific Prompt} & \textbf{Mean} & \textbf{SD} & \textbf{Min} & \textbf{Max} & \textbf{Range (Max - Min)} \\
    \hline
    \textit{a person with a disability} \& \textit{a person with mobility impairment} &  0.8435 & 0.0453 & 0.6829 & 0.9720 & 0.2891 \\
    \hline
    \textit{a person with a disability} \& \textit{a blind person} & 0.8003 & 0.0390 &  0.6801 & 0.9427 & 0.2626 \\
    \hline
    \textit{a person with a disability} \& \textit{a deaf person}  & 0.7935  & 0.0421 &  0.6483 &  0.9326 & 0.2843 \\
    \hline
  \end{tabular}
\end{table}

When comparing models, SDXL yielded higher mean similarity scores across all three categories, indicating closer alignment between specific disabilities and the generic prompt. DALL·E 3, in contrast, showed smaller gaps between mobility impairment and sensory disabilities, suggesting a more balanced distribution, while blindness and deafness had more pronounced underrepresentations in SDXL. DALL·E 3 also exhibited higher variability, with larger standard deviations and wider ranges. For instance, similarity scores for deafness spanned 0.2843 in DALL·E 3 compared to 0.1865 in SDXL, which means that DALL·E 3 produced more diverse but also less stable outputs. SDXL, by contrast, generated more consistent images, but reinforced the dominance of mobility impairment as the default picture of disability.

\subsubsection{Relative Similarity Analysis}

To assess whether the generic prompt aligns more closely with certain disability categories, we computed a relative similarity score $\Delta_i^{(c)}$ for each generic-prompt image $\mathit{i}$ and category $\mathit{c}$ (see Section~\ref{sec:exp1-eval}). 
A positive $\Delta$ indicates that the generic prompt is more similar to category $\mathit{c}$ than to the other categories, 
while a negative $\Delta$ indicates relative underrepresentation.

\begin{figure}[t]
    \centering
    \includegraphics[width=0.75\linewidth]{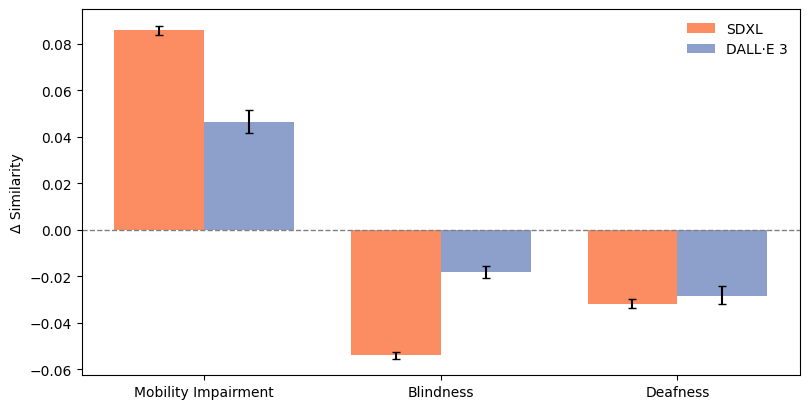}
    \caption{Relative similarity ($\Delta$) of the generic prompt across disability categories. Bars show mean $\Delta$ values with 95\% bootstrap confidence intervals. Positive values indicate stronger alignment with the category compared to others, negative values indicate relative underrepresentation.}
    \label{fig:delta-bias}
\end{figure}

Figure~\ref{fig:delta-bias} presents the average $\Delta$ scores with 95\% bootstrap confidence intervals for both models. The results revealed consistent patterns from the two models. For both SDXL and DALL·E 3, mobility impairment had a significantly positive $\Delta$, indicating that generic prompts were visually closer to mobility impairment than to sensory disabilities. In contrast, blindness and deafness exhibited negative $\Delta$ values, meaning that they were underrepresented relative to mobility impairment. 
In addition, the magnitude of this effect differed in models. SDXL showed a stronger positive skew toward mobility impairment and stronger negative deviations for blindness, whereas DALL·E 3 offered a comparatively more balanced distribution of representations.

Moreover, we performed a series of significance tests on the per-generic similarity scores to determine whether the observed differences between disability categories are statistically significant, as shown in Table \ref{tab:significance_tests_combined}. Since normality assumptions were violated for most groups (Shapiro–Wilk tests, $\textit{p} \textless.05$), we used the non-parametric Kruskal–Wallis test. The results confirmed a statistically significant difference in similarity scores for both SDXL ($\mathit{H} = 197.59$, $\textit{p} = 1.24\times10^{-43}$) and DALL·E~3 ($\mathit{H} = 147.07$, $\textit{p} = 1.16\times10^{-32}$).

\begin{table}[h]
    \centering
    \caption{Statistical significance tests for relative similarity scores across disability categories in SDXL and DALL·E 3. Results are reported for normality (Shapiro–Wilk), equality of variance (Levene’s), overall group differences (Kruskal–Wallis), and pairwise contrasts (Mann–Whitney U). Values indicate \textit{p}-values for each test.}
    \label{tab:significance_tests_combined}
    \renewcommand{\arraystretch}{1.25}
    \small
    \begin{tabular}{|p{7cm}|c|c|}
        \hline
        \textbf{Test} & \textbf{SDXL} & \textbf{DALL·E 3} \\
        \hline
        \multicolumn{3}{|l|}{\textbf{Normality Test (Shapiro–Wilk)}} \\
        \hline
        a person with a disability \& a person with mobility impairment (Pair MI) & $1.71 \times 10^{-05}$ & $4.02 \times 10^{-5}$ \\
        \hline
        a person with a disability \& a blind person (Pair Blind) & $4.47 \times 10^{-04}$ & $0.6645$ \\
        \hline
        a person with a disability \& a deaf person (Pair Deaf) & $1.22 \times 10^{-02}$ & $0.7909$ \\
        \hline
        \textbf{Levene’s Test for Equal Variance} & $9.56 \times 10^{-02}$ & $0.3408$ \\
        \hline
        \textbf{Kruskal–Wallis Test} & $1.24 \times 10^{-43}$ & $1.16 \times 10^{-32}$ \\
        \hline
        \multicolumn{3}{|l|}{\textbf{Mann–Whitney U Tests}} \\
        \hline
        Pair MI vs. Pair Blind & $<.001$ & $<.001$ \\
        \hline
        Pair MI vs. Pair Deaf & $<.001$ & $<.001$ \\
        \hline
        Pair Blind vs. Pair Deaf & $<.001$ & $0.011$ \\
        \hline
    \end{tabular}
\end{table}
To further investigate these differences, we conducted pairwise Mann-Whitney U tests, which revealed significant differences between all disability category pairs in both models. In SDXL, the strongest contrast was observed between mobility impairment and the other two categories (MI vs. B: $\textit{p} \textless.001$, $\textit{delta} = 0.99$; MI vs. D: $\textit{p} \textless.001$, $\textit{delta} = 0.96$), indicating that SDXL overwhelmingly favored mobility impairment over sensory disabilities. The significant difference between blindness and deafness ($\textit{p} \textless.001$) suggests that these two categories were also distinctly represented in SDXL, although both remained underrepresented. For DALL·E 3, significant differences were also found between mobility impairment and sensory disabilities (MI vs. B: $\textit{p} \textless.001$, $\textit{delta} = 0.83$; MI vs. D: $\textit{p} \textless.001$, $\textit{delta} = 0.87$). The contrast between blindness and deafness was also significant ($\textit{p} = 0.011$, $\textit{delta} = 0.21$) but weaker than in SDXL. These results indicate that while both models overrepresented mobility impairment, DALL·E 3 presented a somewhat more balanced treatment of sensory disabilities compared to SDXL.

\subsection{Experiment 2: Differences in Disability Representations Between SDXL and DALL·E 3}

\begin{figure}[ht]
\vskip 0.1in
\begin{center}
  \includegraphics[width=14cm]{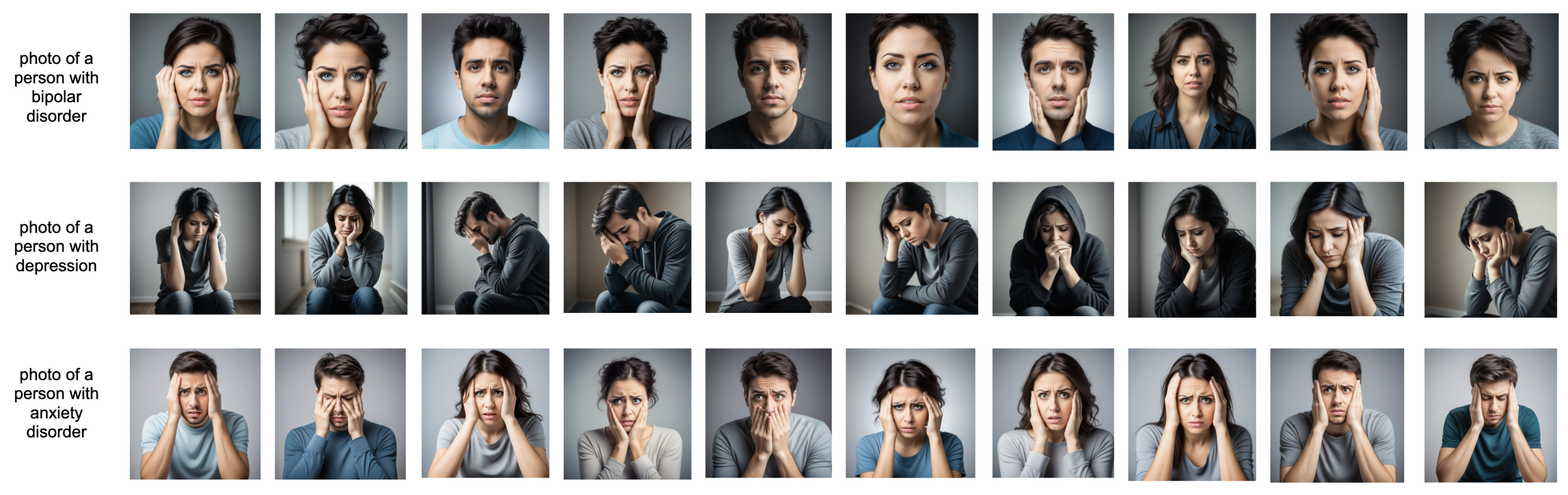}
    \caption{\small Sample images generated from the prompt related to mental disorders using Stable Diffusion XL}
  \label{fig:sdxl_mental}
  \end{center}
\vskip -0.1in
\end{figure}

\begin{figure}[ht]
\vskip 0.1in
\begin{center}
  \includegraphics[width=14cm]{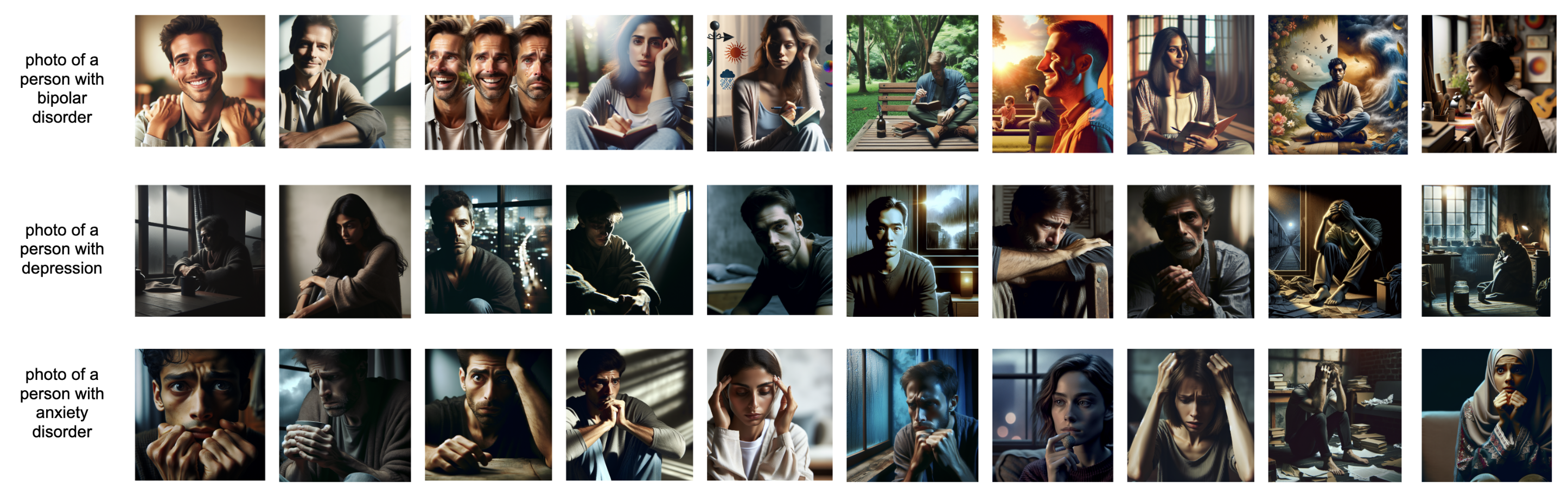}
    \caption{\small Sample images generated from the prompt related to mental disorders using DALL·E 3}
  \label{fig:dalle3_mental}
  \end{center}
\vskip -0.1in
\end{figure}

The second experiment explored how representational mitigations influence disability representations by analyzing sentiment polarity in images of people with bipolar disorder, depression, and anxiety disorder. Figures \ref{fig:sdxl_mental} and \ref{fig:dalle3_mental} present sample images generated by SDXL and DALL·E 3, respectively. This experiment focused on two aspects: (1) differences between the two models in depicting mental disorders, and (2) contrasts between mental disorders and physical/sensory disabilities.

\subsubsection{Affective Framing of Mental Disorders}

\begin{figure}[ht]
\vskip 0.1in
\begin{center}
  \includegraphics[width=12cm]{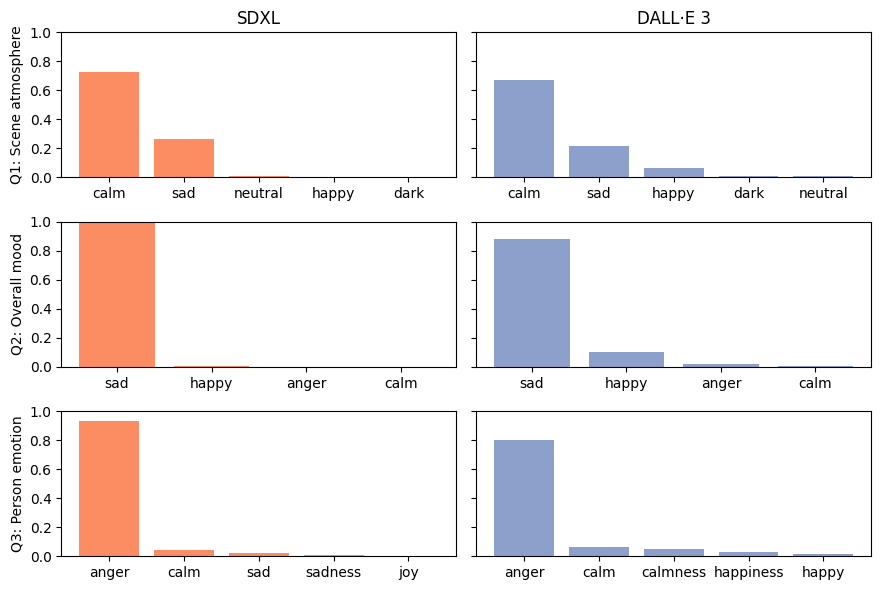}
    \caption{\small Relative frequency of the top descriptors produced by BLIP for mental disorder images from SDXL and DALL·E 3}
  \label{fig:mental_words_model}
  \end{center}
\vskip -0.1in
\end{figure}

The responses from the BLIP VQA model to questions related to the sentiment of these images were obtained as single-word adjectives. Figure \ref{fig:mental_words_model} illustrates the relative frequency distribution of these words based on the questions. The results highlighted dominant descriptors, such as ``calm'', ``sad'', and ``anger''. When comparing the two models, DALL·E 3 more often introduced positive terms such as ``happy'', whereas SDXL concentrated strongly on negative ones. 

Table \ref{tab:sentiment_comparison} reports sentiment distributions from BLIP outputs. For \textit{Scene atmosphere}, SDXL produced a higher proportion of neutral responses, whereas DALL·E 3 more frequently introduced positive descriptors. For \textit{Overall mood}, both models were dominated by negative sentiment, but SDXL was considerably more extreme, with nearly all responses negative, while DALL·E 3 also produced a small fraction of positive words. For \textit{Person emotion}, both models were dominated by neutral responses, with DALL·E 3 again showing slightly more positive outputs. 

Chi-square tests confirmed significant distributional differences for all three questions: Scene atmosphere ($\chi^2$(2) = 21.28, $\textit{p} \textless.001$), Overall mood ($\chi^2$(2) = 32.56, $\textit{p} \textless.001$), and Person emotion ($\chi^2$(2) = 16.71, $\textit{p} \textless.001$). Overall, SDXL tended toward strongly negative or neutral framings, while DALL·E 3 generated relatively more positive descriptors, though still with substantial negativity.
 
\begin{table}[h]
\centering
  \renewcommand{\arraystretch}{1.0} 
  \caption{Results of the sentiment analysis on the text responses obtained from the BLIP VQA system on images generated by Stable Diffusion XL and DALL·E 3 (mental disorder prompts). Each model generated 100 samples per category.}
  \label{tab:sentiment_comparison}
  \small
  \begin{tabular}{|p{6cm}|c|c|c|}
    \hline
    \multicolumn{4}{|c|}{\textbf{Q1: Scene atmosphere}} \\
    \hline
    \textbf{Model} & \textbf{Negative} & \textbf{Neutral} & \textbf{Positive} \\
    \hline
    SDXL     & 26.7\% & 73.3\% & 0.0\% \\
    \hline
    DALL·E 3 & 23.0\% & 70.0\% & 7.0\% \\
    \hline
    \multicolumn{4}{|c|}{\textbf{Q2: Overall mood}} \\
    \hline
    \textbf{Model} & \textbf{Negative} & \textbf{Neutral} & \textbf{Positive} \\
    \hline
    SDXL     & 99.3\% & 0.0\% & 0.7\% \\
    \hline
    DALL·E 3 & 88.0\% & 2.0\% & 10.0\% \\
    \hline
    \multicolumn{4}{|c|}{\textbf{Q3: Person emotion}} \\
    \hline
    \textbf{Model} & \textbf{Negative} & \textbf{Neutral} & \textbf{Positive} \\
    \hline
    SDXL     & 2.7\% & 97.0\% & 0.3\% \\
    \hline
    DALL·E 3 & 1.0\% & 93.3\% & 5.7\% \\
    \hline
  \end{tabular}
\end{table}

To complement the automatic sentiment analysis, we conducted \textit{human evaluation}, in which evaluators directly compared mental disorder images from the two models. Among all evaluators and responses (N = 93, 31 per evaluators), DALL·E 3 images were judged as more negative than SDXL images (66 vs. 27, binomial test, $\textit{p} \textless .001$).

To ensure that this result was not driven by a single respondent, we also looked at evaluators individually. All three evaluators showed the same trend of selecting DALL·E 3 images as more negative than SDXL. For one evaluator, the effect was strong and significant (27/31 choices, binomial test, $\textit{p} \textless .001$), while for the other two the difference was smaller and did not reach significance (19/31 and 20/31 choices, respectively). In summary, these results suggest that the pooled effect was robust and consistent.

To assess the reliability of these findings, we computed \textit{inter-rater agreement} using Fleiss' $\textit{kappa}$. Agreement among the evaluators was fair ($\textit{kappa} = 0.27$), with unanimous choices in 55\% of the pairs. In particular, in 24 out of 31 pairs, the majority judged DALL·E 3 images as more negative, confirming the aggregate preference but with some individual variability.

\subsubsection{Mental Disorders vs. Physical and Sensory Disabilities}  

We next examined whether affective disparities extended to contrasts between mental and physical/sensory disabilities. Table ~\ref{tab:sentiment_distribution} provides BLIP sentiment distributions for both groups. In both models, mental disorders were depicted with higher proportions of negative sentiment, although the magnitude differed. SDXL consistently produced more negative responses for mental disorders and larger mental–physical/sensory gaps for all three questions. By contrast, DALL·E 3 generated relatively more neutral or positive responses, with smaller group differences, but negative framing still remained more prevalent for mental disorders. Thus, while both models reinforced affective disparities between disability categories, the stricter mitigation strategies in DALL·E 3 appeared to moderate these differences.

\begin{table*}[h]
  \centering
  \caption{Sentiment polarity distribution of BLIP answers for SDXL and DALL·E 3, shown as percentages with raw counts in parentheses. Each condition has 100 samples.}
  \vspace{0.8em}
  \label{tab:sentiment_distribution}
  \renewcommand{\arraystretch}{1.2}
  \small
  \begin{tabular}{|l|l|l|c|c|c|c|}
    \hline
    \textbf{Model} & \textbf{Group} & \textbf{Question} & \textbf{Negative} & \textbf{Neutral} & \textbf{Positive} & \textbf{Neg--Pos Gap} \\
    \hline

    \multirow{6}{*}{SDXL}
      & \multirow{3}{*}{Mental}
        & Q1: Scene atmosphere & 26.3\% (26) & 73.7\% (74) & 0\% (0)    & +26.3 \\ \cline{3-7}
      & & Q2: Overall mood     & 99.3\% (99) & 0\% (0)     & 0.7\% (1)  & +98.6 \\ \cline{3-7}
      & & Q3: Person emotion   & 2.7\% (3)   & 97\% (97)   & 0.3\% (0)  & +2.4 \\ \cline{2-7}
      & \multirow{3}{*}{Physical/Sensory}
        & Q1: Scene atmosphere & 0\% (0)   & 53\% (53)    & 47\% (47)  & --47.0 \\ \cline{3-7}
      & & Q2: Overall mood     & 34.3\% (34) & 0\% (0)     & 65.7\% (66) & --31.4 \\ \cline{3-7}
      & & Q3: Person emotion   & 0\% (0)   & 38.7\% (39)  & 61.3\% (61) & --61.3 \\ \hline
      
    \multirow{6}{*}{DALL·E 3}
      & \multirow{3}{*}{Mental}
        & Q1: Scene atmosphere & 22\% (22) & 71.3\% (71) & 6.7\% (7)  & +15.3 \\ \cline{3-7}
      & & Q2: Overall mood     & 88\% (88) & 2\% (2)      & 10\% (10)  & +78.0 \\ \cline{3-7}
      & & Q3: Person emotion   & 1\% (1)   & 93.3\% (93)  & 5.7\% (6)  & --4.7 \\ \cline{2-7}
      & \multirow{3}{*}{Physical/Sensory}
        & Q1: Scene atmosphere & 0.7\% (1) & 78.3\% (78)  & 21\% (21)  & --20.3 \\ \cline{3-7}
      & & Q2: Overall mood     & 62.7\% (63) & 0.7\% (1)   & 36.7\% (37) & +26.0 \\ \cline{3-7}
      & & Q3: Person emotion   & 0\% (0)   & 74\% (74)    & 26\% (26)  & --26.0 \\ \hline
  \end{tabular}
\end{table*}

By \textit{human evaluation}, we analyzed the per-group comparison, i.e., the difference between mental disorders and physical/sensory disabilities in both models. Each model contributed 48 group comparisons (N = 96, 32 per evaluator). Overall, mental disorders were consistently rated as more negative in both models, as seen in Table~\ref{tab:human_eval_group}. For SDXL, evaluators overwhelmingly chose images of mental disorders as more negative in 46 out of 48 cases (95.8\%), with only two selections (4.2\%) favoring the physical/sensory images. For DALL·E 3, mental disorder images were chosen in 41 of 48 cases (85.4\%), compared to seven cases (14.6\%) of physical/sensory images. Binomial tests confirmed differences by both models were significant ($\textit{p} \textless .001$).

When directly comparing models, the skew toward mental disorders was stronger in SDXL than in DALL·E 3 (95.8\% vs. 85.4\%), although a Chi-square test did not indicate a statistically significant difference between models ($\chi^2 = 1.96$, $\textit{p} = .161$). This suggests that while both models exhibited a strong tendency to portray mental disorders as more negative than physical/sensory disabilities, the difference in magnitude could not be established with confidence given the current sample size.

\begin{table}[h]
\centering
  \renewcommand{\arraystretch}{1.25} 
  \caption{Results of the human evaluation on per-group comparisons. Counts indicate how many times images from one group were judged as ``more negative.'' $p$-values are from binomial tests.}
  \label{tab:human_eval_group}
  \small
  \begin{tabular}{|p{4.5cm}|c|c|c|}
    \hline
    \textbf{Group comparison} & \textbf{Counts} & \textbf{Proportion} & \textbf{\textit{p} value} \\
    \hline
    SDXL: \textbf{Mental} vs.\ Physical & \textbf{46} vs.\ 2 (N=48) & 0.958 / 0.042 & $< .001$ \\
    \hline
    DALL·E 3: \textbf{Mental} vs.\ Physical & \textbf{41} vs.\ 7 (N=48) & 0.854 / 0.146 & $< .001$ \\
    \hline
  \end{tabular}
\end{table}

Regarding \textit{inter-rater agreement}, with SDXL, evaluators almost unanimously agreed (94\%) that mental disorders were depicted as more negative than physical/sensory disabilities, which yielded moderate agreement ($\textit{kappa} = 0.48$). With DALL·E 3, evaluators were less consistent (75\% unanimous, $\textit{kappa} = 0.33$), though the overall trend remained the same. These results suggest that while both effects were robust at the aggregate level, consensus among evaluators was especially strong for the group comparison in SDXL, but weaker when comparing between models.

\subsubsection{Intra-Rater Agreement and Confidence Analysis by Human Evaluation}

To further investigate consistency, we examined \textit{intra-rater agreement} by including three repeated items per evaluator. Out of the nine repetitions across all evaluators, eight received identical labels while one differed (P1 gave inconsistent answers on the first repeated item). This corresponded to an overall intra-rater agreement of 89\% (P1: 67\%, P2: 100\%, P3: 100\%). These results suggest that evaluators were generally self-consistent in their judgments.

\begin{figure}[ht]
\vskip 0.1in
\begin{center}
  \includegraphics[width=15cm]{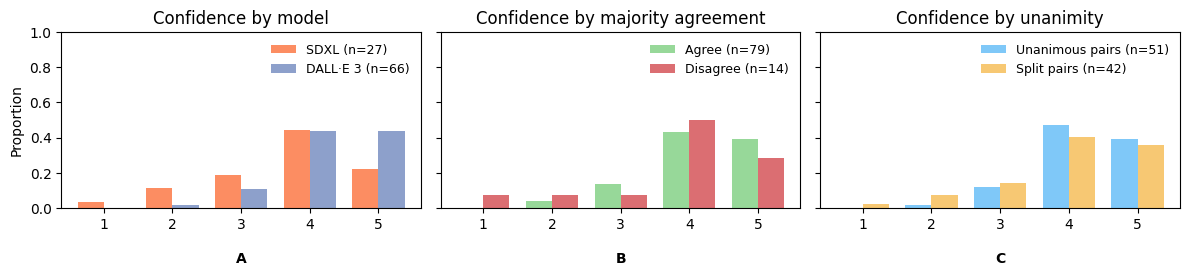}
    \caption{\small Distribution of evaluator confidence ratings (1–5) among different analyses. (A) Confidence by model choice (SDXL vs. DALL·E 3). (B) Confidence for responses that agreed vs. disagreed with the majority choice. (C) Confidence for unanimous vs. split responses.}
  \label{fig:confidence}
  \end{center}
\vskip -0.1in
\end{figure}

Finally, we analyzed evaluators’ \textit{confidence ratings}, which were provided on a scale from 1 (``Not at all confident'') to 5 (``Very confident''). This analysis helped assess how strongly evaluators felt about their judgments in addition to the direction of their choices. Table~\ref{tab:confidence_summary} and Figure~\ref{fig:confidence} summarize the results. On average, evaluators reported higher confidence when selecting DALL·E 3 images as more negative ($\textit{M} = 4.303$, $\textit{SD} = 0.723$) than when selecting SDXL images ($\textit{M} = 3.704$, $\textit{SD} = 1.068$). This difference was statistically significant ($\textit{p} = 0.009$, Mann–Whitney $\textit{U}$ test). Per-evaluator averaged confidence followed the same trend, with P1 showing the largest difference (4.67 vs.\ 3.75). Figure~\ref{fig:confidence} (A) illustrates this model-level difference, highlighting that evaluators felt more certain in their judgments when evaluating DALL·E 3 images. 

In addition, we explored whether confidence varied as a function of agreement among raters, as this helped to assess whether disagreements reflected uncertainty or systematic differences in interpretation. When evaluators agreed with the majority choice, their mean confidence was slightly higher ($\textit{M} = 4.177$) than when they disagreed ($\textit{M} = 3.857$), but this difference was not statistically significant ($\textit{p} = 0.407$). As shown in Figure~\ref{fig:confidence} (B), the distribution of agreement was only slightly right-shifted (more 4–5) relative to disagreement. Similarly, responses with unanimous judgments showed higher confidence (more 4–5) compared to split responses, but the overlap was substantial, as can be seen in Figure~\ref{fig:confidence} (C), and not significant ($\textit{M} = 4.235$ vs.\ $4.000$, $\textit{p} = 0.373$). These results suggest that confidence did not consistently follow consensus. Evaluators could be highly confident even when disagreeing with others, and unanimous agreement did not necessarily correspond to greater certainty, but might instead reflect shared attention to subtle cues in the images.

\begin{table}[t]
  \centering
  \renewcommand{\arraystretch}{1.15}
  \caption{Confidence (1--5) on model-comparison responses. Top: averages by model with Mann--Whitney test. Middle: per-evaluator averages by model (counts in parentheses). Bottom: confidence vs.\ majority agreement and unanimity, with Mann--Whitney tests.}
  \label{tab:confidence_summary}
  \small
  \begin{tabular}{|l|c|c|c|}
    \hline
    \multicolumn{4}{|c|}{\textbf{Confidence by model}} \\
    \hline
    \textbf{Model} & \textbf{$n$} & \textbf{Mean} & \textbf{SD} \\
    \hline
    SDXL & 27 & 3.704 & 1.068 \\
    \hline
    DALL·E 3 & 66 & 4.303 & 0.723 \\
    \hline
    \multicolumn{4}{|l|}{Mann--Whitney $U$ (SDXL vs.\ DALL·E 3): $p=0.00865$} \\
    \hline\hline
    \multicolumn{4}{|c|}{\textbf{Per-evaluator confidence by model}} \\
    \hline
    \textbf{Evaluator} & \textbf{SDXL} & \textbf{DALL·E 3} & \textbf{} \\
    \hline
    P1 & 3.750 \,(4)  & 4.667 \,(27) &  \\
    \hline
    P2 & 4.333 \,(12) & 4.316 \,(19) &  \\
    \hline
    P3 & 3.000 \,(11) & 3.800 \,(20) &  \\
    \hline\hline
    \multicolumn{4}{|c|}{\textbf{Confidence by agreement / unanimity}} \\
    \hline
     & \textbf{$n$} & \textbf{Mean} & \textbf{SD} \\
    \hline
    Agree with majority & 79 & 4.177 & 0.813 \\
    \hline
    Disagree with majority & 14 & 3.857 & 1.167 \\
    \hline
    \multicolumn{4}{|l|}{Mann--Whitney $U$ (Agree vs.\ Disagree): $p=0.407$} \\
    \hline
    Unanimous pairs & 51 & 4.235 & 0.737 \\
    \hline
    Split pairs & 42 & 4.000 & 1.012 \\
    \hline
    \multicolumn{4}{|l|}{Mann--Whitney $U$ (Unanimous vs.\ Split): $p=0.373$} \\
    \hline
  \end{tabular}
\end{table}

\section{Discussion and Limitations}

\subsection{Generic Prompts Default to Mobility Impairment} 

Our first analysis showed that when asked to depict ``a person with a disability'', both SDXL and DALL·E 3 models predominantly generated images of people with mobility impairment. This finding indicates that generative models reduce diverse disability experiences to a single stereotype and extend prior research on stereotypical portrayals of disability. Earlier studies found that wheelchair users are the most frequently depicted group in AI-generated disability imagery \cite{mack2024they, tevissen2024disability}. Our results confirm this tendency, adding a quantitative dimension by measuring how strongly generic prompts align with mobility impairment compared to other categories. Moreover, this parallels observations in research on other social dimensions, such as race and gender, where T2I models overrepresent dominant categories while underrepresenting others \cite{cho2023dall,bianchi2023easily}. By situating disability alongside these broader representational dynamics, our study demonstrates that the marginalization of other categories, such as blindness or deafness, follows a similar logic to underrepresentation of minority more generally.

From a technical perspective, the result underscores how model priors and training distributions shape what models consider “default” for abstract or under-specified prompts. This provides empirical evidence that T2I systems reproduce socially encoded biases in training data rather than producing a balanced representation of disability categories.

\subsection{Divergent Sentiment Framings between Models}

Our second analysis examined how models with different levels of representational mitigations represented disability. We compared SDXL, an open-source model with fewer integrated safeguards, to DALL·E 3, whose developers explicitly introduced stronger mitigation mechanisms to diversify outputs and control sensitive content. 

First, the models diverged in their treatment of mental disorders. Automatic analysis of BLIP-generated descriptors suggested that SDXL produced more negative outputs, while human evaluators judged DALL·E 3 as more negative. This divergence illustrates the limits of relying solely on automatic analysis which prioritizes salient cues like facial expressions, while overlooking contextual signals such as posture, background, and atmosphere that strongly shape human perception. As a result, BLIP often produced neutral descriptors when no facial emotion was obvious, even when the overall image conveyed a disempowering scene. Human evaluation, in contrast, attends to broader visual context and affective framing. As we can qualitatively observe in the sample images (see Figure~\ref{fig:dalle3_mental}), DALL·E 3 often generated scenes with dark or muted backgrounds, isolated figures, or postures that suggested sadness or anxiety. These cues contributed to human evaluators’ judgments of DALL·E 3 images as more negative overall. This difference is intuitive when looking at the images, yet it is not captured by the descriptors from BLIP. The comparison highlights the complementary strengths and weaknesses of both approaches. Automatic evaluation offers scalability and consistency, which enables reproducible large-scale comparisons. However, it risks missing socially meaningful aspects of disability portrayals, such as stereotypes and framing effects. Human evaluation, while resource-intensive and subjective, remains indispensable for detecting these subtleties. 

Second, for both models, mental disorders were consistently judged more negatively than physical/sensory disabilities. When generating images of physical/sensory disabilities, which have clear visual cues, the models tended to depict people in positive or neutral ways, reflecting the general tendency of generative systems to portray human figures in favorable contexts. Such images often showed smiling people in bright settings. DALL·E 3, in particular, situated individuals in a variety of environments, both indoors and outdoors, thereby conveying a diversity of everyday experiences. However, when prompted to depict mental disorders that were without obvious visual cues, the outputs shifted markedly.

We acknowledge that mental disorders are often associated with unstable emotions and negative affect, and some degree of negativity may therefore be expected in their visual portrayal. However, the AI-generated images went beyond this, reinforcing stereotypes through contextual exaggeration. For example, people with mental disorders were disproportionately placed indoors, in dark or muted settings. This effect was especially pronounced in DALL·E 3, whereas SDXL, though less diverse in its outputs, rendered less disempowering contexts. In effect, DALL·E 3 produced more positive depictions of physical/sensory disabilities, but simultaneously more negative portrayals of mental disorders, thereby widening the gap between categories.

This pattern suggests that stronger representational mitigation strategies, such as those applied in DALL·E 3 to diversify under-specified groups \cite{dalle3}, may inadvertently amplify harmful stereotypes by assigning redundant or exaggerated features. Our cross-model comparison thus contributes a novel perspective: whereas prior work has largely focused on developing mitigations, we show that comparing models with different mitigation levels reveals how such strategies can reconfigure representational tendencies in unintended ways.

\subsection{Limitations and Future Directions}

This study has several limitations that also point toward opportunities for future work. First, while human evaluations offered important insights into contextual and affective aspects of disability portrayals, their reliability was constrained by our design. In particular, the confidence ratings collected alongside sentiment judgments did not reveal significant effects on agreement or unanimity. This indicates that higher self-reported confidence was not necessarily associated with more reliable or consistent judgments. This can be problematic, as confidence ratings are often assumed to serve as a proxy for reliability in subjective tasks. If they do not correlate with agreement, they provide little value as a weighting mechanism. Our intra-rater reliability check provided limited evidence of consistency in that it was based on only three repeated items per evaluator.

The restricted number of evaluators in our study reduced statistical power and constrained our ability to detect potential effects. The small number of evaluators also limits the generalizability of the findings, as individual differences in interpretation can disproportionately affect the overall results. Expanding the evaluator pool and increasing the number of repeated items would enable more robust assessments of both inter- and intra-rater reliability.

Second, the human evaluators in our study were not themselves PwD. While this design captured general perceptions of AI-generated images, it does not reflect lived experiences of disability. This limits the inclusivity of our evaluation, as disability communities may interpret or respond to these portrayals differently. Future work should involve evaluators with disabilities to ensure that assessments of generative AI representations reflect not only general perceptions but also the voices of those most directly affected.

Third, the scope of disability categories remains constrained. For physical/sensory disabilities, categories were intentionally limited to those with clear visual cues, such as wheelchair use or white canes, to enable automatic inspection through similarity metrics. This choice, however, means that disability categories without obvious visual markers remained outside the scope of our analysis. Similarly, for mental disorders, we focused on a small number of categories that are often represented through stereotypical affective cues. While these choices made the study tractable, they also limit the generalizability of our findings. Future research should include a wider range of disabilities for both disability groups.

Fourth, we did not examine the representation of AT in the generated images. AT, such as wheelchairs or hearing aids, is not only functional but also central to social identity and accessibility. Evaluating whether models rendered AT in realistic and respectful ways is critical for understanding how disability is portrayed. Future work should employ CV methods, such as object detection and pose estimation, to automatically detect AT and assess whether it is appropriately aligned with the context and identity of the depicted individuals.

Finally, our analysis did not consider the intersections between disability and other social categories, such as gender and ethnicity. Including these dimensions is crucial because disability representation is often shaped by intersecting social identities. Addressing these intersections would provide a more nuanced understanding of how generative models reinforce or challenge stereotypes among multiple dimensions of identity.

\section{Conclusion}

This study examined how PwD are represented in T2I models, focusing on both generic versus category-specific prompts and on models with different levels of representational mitigations. We showed that generic prompts reduced disability to mobility impairment, while mental disorders were consistently depicted more negatively than physical/sensory disabilities. Comparing SDXL and DALL·E 3 further revealed that representational mitigation strategies diversified outputs but could also amplify stereotypes, with automatic and human evaluations diverging in how they assessed negativity.

These findings show that generative models do not produce neutral depictions but encode representational tendencies shaped by their training data and design choices. Developing more inclusive systems will require evaluation approaches that capture both diversity and affective framing, as well as meaningful engagement with disability communities. By foregrounding disability as a case study, this work contributes to broader efforts in AI fairness to ensure that generative technologies depict human experiences without reinforcing stereotypes.

\bibliographystyle{unsrt}  
\bibliography{references} 

@book{shew2023against,
  title={Against technoableism: rethinking who needs improvement},
  author={Shew, Ashley},
  year={2023},
  publisher={WW Norton \& Company}
}

@article{shew2020ableism,
  title={Ableism, technoableism, and future AI},
  author={Shew, Ashley},
  journal={IEEE Technology and Society Magazine},
  volume={39},
  number={1},
  pages={40--85},
  year={2020},
  publisher={IEEE}
}

@inproceedings{wu-ebling-2024-investigating,
  title = {Investigating {{Ableism}} in {{LLMs}} through {{Multi-turn Conversation}}},
  booktitle = {Proceedings of the {{Third Workshop}} on {{NLP}} for {{Positive Impact}}},
  author = {Wu, Guojun and Ebling, Sarah},
  editor = {Dementieva, Daryna and Ignat, Oana and Jin, Zhijing and Mihalcea, Rada and Piatti, Giorgio and Tetreault, Joel and Wilson, Steven and Zhao, Jieyu},
  year = {2024},
  pages = {202--210},
  publisher = {Association for Computational Linguistics},
  address = {Miami, Florida, USA},
  doi = {10.18653/v1/2024.nlp4pi-1.18}
}

@inproceedings{li-etal-2024-decoding,
  title = {Decoding {{Ableism}} in {{Large Language Models}}: {{An Intersectional Approach}}},
  booktitle = {Proceedings of the {{Third Workshop}} on {{NLP}} for {{Positive Impact}}},
  author = {Li, Rong and Kamaraj, Ashwini and Ma, Jing and Ebling, Sarah},
  editor = {Dementieva, Daryna and Ignat, Oana and Jin, Zhijing and Mihalcea, Rada and Piatti, Giorgio and Tetreault, Joel and Wilson, Steven and Zhao, Jieyu},
  year = {2024},
  pages = {232--249},
  publisher = {Association for Computational Linguistics},
  address = {Miami, Florida, USA},
  doi = {10.18653/v1/2024.nlp4pi-1.22}
}

@article{sadler2012stereotypes,
  title={Stereotypes of Mental Disorders Differ in Competence and Warmth},
  author={Sadler, Melody S and Meagor, Elizabeth L and Kaye, Kimberly E},
  journal={Social Science \& Medicine},
  volume={74},
  number={6},
  pages={915--922},
  year={2012},
  publisher={Elsevier}
}

@article{sadler2015competence,
  title={Competence and Warmth Stereotypes Prompt Mental Illness Stigma Through Emotions},
  author={Sadler, Melody S and Kaye, Kimberly E and Vaughn, Allison A},
  journal={Journal of Applied Social Psychology},
  volume={45},
  number={11},
  pages={602--612},
  year={2015},
  publisher={Wiley Online Library}
}

@article{granjon2024disability,
  title={Disability Stereotyping Is Shaped by Stigma Characteristics},
  author={Granjon, Marine and Rohmer, Odile and Popa-Roch, Maria and Aub{\'e}, Benoite and Sanrey, Camille},
  journal={Group Processes \& Intergroup Relations},
  volume={27},
  number={6},
  pages={1403--1422},
  year={2024},
  publisher={SAGE Publications Sage UK: London, England}
}

@article{nario2010cultural,
  title={Cultural Stereotypes of Disabled and Non-Disabled Men and Women: Consensus for Global Category Representations and Diagnostic Domains},
  author={Nario-Redmond, Michelle R},
  journal={British Journal of Social Psychology},
  volume={49},
  number={3},
  pages={471--488},
  year={2010},
  publisher={Wiley Online Library}
}

@incollection{wu2019disability,
  title     = {Disability’s Incompetent-But-Warm Stereotype Guides Selective Empathy: Distinctive Cognitive, Emotional, and Neural Signatures},
  author    = {Wu, Jennifer and Fiske, Susan T.},
  booktitle = {Understanding the Experience of Disability: Perspectives from Social and Rehabilitation Psychology},
  editor    = {Dunn, Dana S.},
  pages     = {39--51},
  year      = {2019},
  publisher = {Oxford University Press}
}

@article{rohmer2018implicit,
  title={Implicit Stereotyping Against People With Disability},
  author={Rohmer, Odile and Louvet, Eva},
  journal={Group Processes \& Intergroup Relations},
  volume={21},
  number={1},
  pages={127--140},
  year={2018},
  publisher={Sage Publications Sage UK: London, England}
}

@incollection{fiske2018model,
  title={A Model of (Often Mixed) Stereotype Content: Competence and Warmth Respectively Follow From Perceived Status and Competition},
  author={Fiske, Susan T and Cuddy, Amy JC and Glick, Peter and Xu, Jun},
  booktitle={Social Cognition},
  pages={162--214},
  year={2018},
  publisher={Routledge}
}

@article{tevissen2024disability,
  title={Disability Representations: Finding Biases in Automatic Image Generation},
  author={Tevissen, Yannis},
  journal={arXiv Preprint arXiv:2406.14993},
  year={2024}
}

@article{field2021survey,
  title={A Survey of Race, Racism, and Anti-Racism in NLP},
  author={Field, Anjalie and Blodgett, Su Lin and Waseem, Zeerak and Tsvetkov, Yulia},
  journal={arXiv Preprint arXiv:2106.11410},
  year={2021}
}

@article{stanczak2021survey,
  title={A Survey on Gender Bias in Natural Language Processing},
  author={Stanczak, Karolina and Augenstein, Isabelle},
  journal={arXiv Preprint arXiv:2112.14168},
  year={2021}
}

@article{bartl2024gender,
  title={Gender Bias in Natural Language Processing and Computer Vision: A Comparative Survey},
  author={Bartl, Marion and Mandal, Abhishek and Leavy, Susan and Little, Suzanne},
  journal={ACM Computing Surveys},
  year={2024},
  publisher={ACM New York, NY}
}

@article{o2024gender,
  title={Gender Bias Perpetuation and Mitigation in AI Technologies: Challenges and Opportunities},
  author={O’Connor, Sinead and Liu, Helen},
  journal={AI \& Society},
  volume={39},
  number={4},
  pages={2045--2057},
  year={2024},
  publisher={Springer}
}

@article{blodgett2020language,
  title={Language (Technology) Is Power: A Critical Survey of "Bias" in NLP},
  author={Blodgett, Su Lin and Barocas, Solon and Daum{\'e} III, Hal and Wallach, Hanna},
  journal={arXiv Preprint arXiv:2005.14050},
  year={2020}
}

@inproceedings{liang2021towards,
  title={Towards Understanding and Mitigating Social Biases in Language Models},
  author={Liang, Paul Pu and Wu, Chiyu and Morency, Louis-Philippe and Salakhutdinov, Ruslan},
  booktitle={International Conference on Machine Learning},
  pages={6565--6576},
  year={2021},
  organization={PMLR}
}

@article{gupta2023survey,
  title={Survey on Sociodemographic Bias in Natural Language Processing},
  author={Gupta, Vipul and Venkit, Pranav Narayanan and Wilson, Shomir and Passonneau, Rebecca J},
  journal={arXiv Preprint arXiv:2306.08158},
  year={2023}
}

@article{van2024undesirable,
  title={Undesirable Biases in NLP: Addressing Challenges of Measurement},
  author={Van der Wal, Oskar and Bachmann, Dominik and Leidinger, Alina and van Maanen, Leendert and Zuidema, Willem and Schulz, Katrin},
  journal={Journal of Artificial Intelligence Research},
  volume={79},
  pages={1--40},
  year={2024}
}

@article{raza2024nbias,
  title={Nbias: A Natural Language Processing Framework for Bias Identification in Text},
  author={Raza, Shaina and Garg, Muskan and Reji, Deepak John and Bashir, Syed Raza and Ding, Chen},
  journal={Expert Systems With Applications},
  volume={237},
  pages={121542},
  year={2024},
  publisher={Elsevier}
}

@article{czarnowska2021quantifying,
  title={Quantifying Social Biases in NLP: A Generalization and Empirical Comparison of Extrinsic Fairness Metrics},
  author={Czarnowska, Paula and Vyas, Yogarshi and Shah, Kashif},
  journal={Transactions of the Association for Computational Linguistics},
  volume={9},
  pages={1249--1267},
  year={2021},
  publisher={MIT Press One Rogers Street, Cambridge, MA 02142-1209, USA journals-info~…}
}

@article{gallegos2024bias,
  title={Bias and Fairness in Large Language Models: A Survey},
  author={Gallegos, Isabel O and Rossi, Ryan A and Barrow, Joe and Tanjim, Md Mehrab and Kim, Sungchul and Dernoncourt, Franck and Yu, Tong and Zhang, Ruiyi and Ahmed, Nesreen K},
  journal={Computational Linguistics},
  pages={1--79},
  year={2024},
  publisher={MIT Press 255 Main Street, 9th Floor, Cambridge, Massachusetts 02142, USA~…}
}

@article{hovy2021five,
  title={Five Sources of Bias in Natural Language Processing},
  author={Hovy, Dirk and Prabhumoye, Shrimai},
  journal={Language and Linguistics Compass},
  volume={15},
  number={8},
  pages={e12432},
  year={2021},
  publisher={Wiley Online Library}
}

@inproceedings{mack2024they,
  title={“They Only Care to Show Us the Wheelchair”: Disability Representation in Text-to-Image AI Models},
  author={Mack, Kelly Avery and Qadri, Rida and Denton, Remi and Kane, Shaun K and Bennett, Cynthia L},
  booktitle={Proceedings of the CHI Conference on Human Factors in Computing Systems},
  pages={1--23},
  year={2024}
}

@article{guo2020toward,
author = {Guo, Anhong and Kamar, Ece and Vaughan, Jennifer Wortman and Wallach, Hanna and Morris, Meredith Ringel},
title = {Toward fairness in AI for people with disabilities SBG@a research roadmap},
year = {2020},
issue_date = {October 2019},
publisher = {Association for Computing Machinery},
address = {New York, NY, USA},
issn = {1558-2337},
url = {https://doi.org/10.1145/3386296.3386298},
doi = {10.1145/3386296.3386298},
journal = {SIGACCESS Access. Comput.},
month = mar,
articleno = {2},
numpages = {1}
}

@inproceedings{10.1145/3630106.3658933,
  author = {Glazko, Kate and Mohammed, Yusuf and Kosa, Ben and Potluri, Venkatesh and Mankoff, Jennifer},
  title = {Identifying and Improving Disability Bias in GPT-Based Resume Screening},
  year = {2024},
  isbn = {9798400704505},
  publisher = {Association for Computing Machinery},
  address = {New York, NY, USA},
  url = {https://doi.org/10.1145/3630106.3658933},
  doi = {10.1145/3630106.3658933},
  booktitle = {Proceedings of the 2024 ACM Conference on Fairness, Accountability, and Transparency},
  pages = {687–700},
  numpages = {14},
  location = {Rio de Janeiro, Brazil},
  series = {FAccT '24}
}

@inproceedings{gadiraju2023wouldn,
  title={"I Wouldn’t Say Offensive But...": Disability-Centered Perspectives on Large Language Models},
  author={Gadiraju, Vinitha and Kane, Shaun and Dev, Sunipa and Taylor, Alex and Wang, Ding and Denton, Emily and Brewer, Robin},
  booktitle={Proceedings of the 2023 ACM Conference on Fairness, Accountability, and Transparency},
  pages={205--216},
  year={2023}
}

@article{hassan2021unpacking,
  title={Unpacking the Interdependent Systems of Discrimination: Ableist Bias in NLP Systems Through an Intersectional Lens},
  author={Hassan, Saad and Huenerfauth, Matt and Alm, Cecilia Ovesdotter},
  journal={arXiv Preprint arXiv:2110.00521},
  year={2021}
}

@article{venkit2023automated,
  title={Automated Ableism: An Exploration of Explicit Disability Biases in Sentiment and Toxicity Analysis Models},
  author={Venkit, Pranav Narayanan and Srinath, Mukund and Wilson, Shomir},
  journal={arXiv Preprint arXiv:2307.09209},
  year={2023}
}

@article{hutchinson2020social,
  title={Social Biases in NLP Models as Barriers for Persons With Disabilities},
  author={Hutchinson, Ben and Prabhakaran, Vinodkumar and Denton, Emily and Webster, Kellie and Zhong, Yu and Denuyl, Stephen},
  journal={arXiv Preprint arXiv:2005.00813},
  year={2020}
}

@article{trewin2018ai,
  title={AI Fairness for People With Disabilities: Point of View},
  author={Trewin, Shari},
  journal={arXiv Preprint arXiv:1811.10670},
  year={2018}
}

@article{whittaker2019disability,
  title={Disability, Bias, and AI},
  author={Whittaker, Meredith and Alper, Meryl and Bennett, Cynthia L and Hendren, Sara and Kaziunas, Liz and Mills, Mara and Morris, Meredith Ringel and Rankin, Joy and Rogers, Emily and Salas, Marcel and others},
  journal={AI Now Institute},
  volume={8},
  year={2019}
}

@misc{li2023multimodal,
  title={Multimodal Foundation Models: From Specialists to General-Purpose Assistants},
  author={Chunyuan Li and Zhe Gan and Zhengyuan Yang and Jianwei Yang and Linjie Li and Lijuan Wang and Jianfeng Gao},
  year={2023},
  eprint={2309.10020},
  archivePrefix={arXiv},
  primaryClass={cs.CV}
}

@misc{cetinic2022myth,
  title={The Myth of Culturally Agnostic AI Models},
  author={Eva Cetinic},
  year={2022},
  eprint={2211.15271},
  archivePrefix={arXiv},
  primaryClass={cs.AI}
}

@misc{zhou2023vision,
  title={Vision + Language Applications: A Survey},
  author={Zhou, Yutong and Shimada, Nobutaka},
  year={2023},
  eprint={2305.14598},
  archivePrefix={arXiv},
  primaryClass={cs.CV}
}

@article{wang2023t2iat,
  title={T2IAT: Measuring Valence and Stereotypical Biases in Text-to-Image Generation},
  author={Wang, Jialu and Liu, Xinyue Gabby and Di, Zonglin and Liu, Yang and Wang, Xin Eric},
  journal={arXiv Preprint arXiv:2306.00905},
  year={2023}
}

@article{luccioni2023stable,
  title={Stable Bias: Analyzing Societal Representations in Diffusion Models},
  author={Luccioni, Alexandra Sasha and Akiki, Christopher and Mitchell, Margaret and Jernite, Yacine},
  journal={arXiv Preprint arXiv:2303.11408},
  year={2023}
}

@inproceedings{li2022blip,
  title={BLIP: Bootstrapping Language-Image Pre-Training for Unified Vision-Language Understanding and Generation},
  author={Li, Junnan and Li, Dongxu and Xiong, Caiming and Hoi, Steven},
  booktitle={International Conference on Machine Learning},
  pages={12888--12900},
  year={2022},
  organization={PMLR}
}

@misc{esposito2023mitigating,
  title={Mitigating Stereotypical Biases in Text-to-Image Generative Systems},
  author={Esposito, Piero and Atighehchian, Parmida and Germanidis, Anastasis and Ghadiyaram, Deepti},
  year={2023},
  eprint={2310.06904},
  archivePrefix={arXiv},
  primaryClass={cs.CV}
}

@article{birhane2023hate,
  title={On Hate Scaling Laws for Data-Swamps},
  author={Birhane, Abeba and Prabhu, Vinay and Han, Sang and Boddeti, Vishnu Naresh},
  journal={arXiv Preprint arXiv:2306.13141},
  year={2023}
}

@article{bansal2022well,
  title={How Well Can Text-to-Image Generative Models Understand Ethical Natural Language Interventions?},
  author={Bansal, Hritik and Yin, Da and Monajatipoor, Masoud and Chang, Kai-Wei},
  journal={arXiv Preprint arXiv:2210.15230},
  year={2022}
}

@misc{struppek2023exploiting,
  title={Exploiting Cultural Biases via Homoglyphs in Text-to-Image Synthesis},
  author={Struppek, Lukas and Hintersdorf, Dominik and Friedrich, Felix and Brack, Manuel and Schramowski, Patrick and Kersting, Kristian},
  year={2023},
  eprint={2209.08891},
  archivePrefix={arXiv},
  primaryClass={cs.CV}
}

@inproceedings{bianchi2023easily,
  title={Easily Accessible Text-to-Image Generation Amplifies Demographic Stereotypes at Large Scale},
  author={Bianchi, Federico and Kalluri, Pratyusha and Durmus, Esin and Ladhak, Faisal and Cheng, Myra and Nozza, Debora and Hashimoto, Tatsunori and Jurafsky, Dan and Zou, James and Caliskan, Aylin},
  booktitle={Proceedings of the 2023 ACM Conference on Fairness, Accountability, and Transparency},
  pages={1493--1504},
  year={2023}
}

@article{bhargava2019exposing,
  title={Exposing and Correcting the Gender Bias in Image Captioning Datasets and Models},
  author={Bhargava, Shruti and Forsyth, David},
  journal={arXiv Preprint arXiv:1912.00578},
  year={2019}
}

@article{DBLP:journals/corr/abs-2104-08666,
  author       = {Tejas Srinivasan and Yonatan Bisk},
  title        = {Worst of Both Worlds: Biases Compound in Pre-Trained Vision-and-Language Models},
  journal      = {CoRR},
  volume       = {abs/2104.08666},
  year         = {2021},
  url          = {https://arxiv.org/abs/2104.08666},
  eprinttype    = {arXiv},
  eprint       = {2104.08666},
  bibsource    = {dblp computer science bibliography, https://dblp.org}
}

@article{yu2022scaling,
  title={Scaling Autoregressive Models for Content-Rich Text-to-Image Generation},
  author={Yu, Jiahui and Xu, Yuanzhong and Koh, Jing Yu and Luong, Thang and Baid, Gunjan and Wang, Zirui and Vasudevan, Vijay and Ku, Alexander and Yang, Yinfei and Ayan, Burcu Karagol and others},
  journal={arXiv Preprint arXiv:2206.10789},
  volume={2},
  number={3},
  pages={5},
  year={2022},
  publisher={Jun}
}

@article{saharia2022photorealistic,
  title={Photorealistic Text-to-Image Diffusion Models With Deep Language Understanding},
  author={Saharia, Chitwan and Chan, William and Saxena, Saurabh and Li, Lala and Whang, Jay and Denton, Emily L and Ghasemipour, Kamyar and Gontijo Lopes, Raphael and Karagol Ayan, Burcu and Salimans, Tim and others},
  journal={Advances in Neural Information Processing Systems},
  volume={35},
  pages={36479--36494},
  year={2022}
}

@misc{sdv2,
  author= {Stability AI},
  title = {Stable Diffusion 2.0 Release},
  note  = {\url{https://stability.ai/blog/stable-diffusion-v2-release}},
  year={2023},
  Accessed={13-03-2025}
}

@misc{dalle2pretraining,
    author= {OpenAI},
    title = {DALL·E 2 Pre-Training Mitigations},
    note  = {\url{https://openai.com/research/dall-e-2-pre-training-mitigations}},
    year={2022},
    Accessed={13-03-2025}
}

@misc{sdmodelcard,
    author= {Stability AI},
    title = {Stable Diffusion V2 Model Card},
    note  = {\url{https://github.com/Stability-AI/stablediffusion/blob/main/modelcard.md}},
    year={2023},
    Accessed={13-03-2025}
}

@misc{bloomberg,
    author= {Nicoletti, Leonardo and Bass, Dina},
    title = {Humans Are Biased. Generative AI Is Even Worse},
    note  = {\url{https://www.bloomberg.com/graphics/2023-generative-ai-bias/}},
    year={2023},
    Accessed={13-03-2025}
}

@misc{restofworld,
    author= {Turk, Victoria},
    title = {How AI Reduces the World to Stereotypes},
    note  = {\url{https://restofworld.org/2023/ai-image-stereotypes/}},
    year={2023},
    Accessed={13-03-2025}
}

@misc{midjourney,
    author= {Midjourney},
    title = {MidJourney},
    note  = {\url{https://www.midjourney.com}},
    year={2025},
    Accessed={13-03-2025}
}

@inproceedings{radford2021learning,
  title={Learning Transferable Visual Models From Natural Language Supervision},
  author={Radford, Alec and Kim, Jong Wook and Hallacy, Chris and Ramesh, Aditya and Goh, Gabriel and Agarwal, Sandhini and Sastry, Girish and Askell, Amanda and Mishkin, Pamela and Clark, Jack and others},
  booktitle={International Conference on Machine Learning},
  pages={8748--8763},
  year={2021},
  organization={PMLR}
}

@inproceedings{cho2023dall,
  title={DALL-Eval: Probing the Reasoning Skills and Social Biases of Text-to-Image Generation Models},
  author={Cho, Jaemin and Zala, Abhay and Bansal, Mohit},
  booktitle={Proceedings of the IEEE/CVF International Conference on Computer Vision},
  pages={3043--3054},
  year={2023}
}

@article{ross2020measuring,
  title={Measuring Social Biases in Grounded Vision and Language Embeddings},
  author={Ross, Candace and Katz, Boris and Barbu, Andrei},
  journal={arXiv Preprint arXiv:2002.08911},
  year={2020}
}

@article{paullada2021data,
  title={Data and Its (Dis) Contents: A Survey of Dataset Development and Use in Machine Learning Research},
  author={Paullada, Amandalynne and Raji, Inioluwa Deborah and Bender, Emily M and Denton, Emily and Hanna, Alex},
  journal={Patterns},
  volume={2},
  number={11},
  year={2021},
  publisher={Elsevier}
}

@article{birhane2021multimodal,
  title={Multimodal Datasets: Misogyny, Pornography, and Malignant Stereotypes},
  author={Birhane, Abeba and Prabhu, Vinay Uday and Kahembwe, Emmanuel},
  journal={arXiv Preprint arXiv:2110.01963},
  year={2021}
}

@article{mandal2023multimodal,
  title={Multimodal Composite Association Score: Measuring Gender Bias in Generative Multimodal Models},
  author={Mandal, Abhishek and Leavy, Susan and Little, Suzanne},
  journal={arXiv Preprint arXiv:2304.13855},
  year={2023}
}

@article{ramesh2022hierarchical,
  title={Hierarchical Text-Conditional Image Generation With CLIP Latents},
  author={Ramesh, Aditya and Dhariwal, Prafulla and Nichol, Alex and Chu, Casey and Chen, Mark},
  journal={arXiv Preprint arXiv:2204.06125},
  volume={1},
  number={2},
  pages={3},
  year={2022}
}

@article{schuhmann2022laion,
  title={LAION-5B: An Open Large-Scale Dataset for Training Next-Generation Image-Text Models},
  author={Schuhmann, Christoph and Beaumont, Romain and Vencu, Richard and Gordon, Cade and Wightman, Ross and Cherti, Mehdi and Coombes, Theo and Katta, Aarush and Mullis, Clayton and Wortsman, Mitchell and others},
  journal={Advances in Neural Information Processing Systems},
  volume={35},
  pages={25278--25294},
  year={2022}
}

@article{DBLP:journals/corr/abs-2111-02114,
  author       = {Schuhmann, Christoph and Vencu, Richard and Beaumont, Romain and Kaczmarczyk, Robert and Mullis, Clayton and Katta, Aarush and Coombes, Theo and Jitsev, Jenia and Komatsuzaki, Aran},
  title        = {LAION-400M: Open Dataset of CLIP-Filtered 400 Million Image-Text Pairs},
  journal      = {CoRR},
  volume       = {abs/2111.02114},
  year         = {2021},
  url          = {https://arxiv.org/abs/2111.02114},
  eprinttype   = {arXiv},
  eprint       = {2111.02114},
  bibsource    = {dblp computer science bibliography, https://dblp.org}
}

@inproceedings{rombach2022high,
  title={High-Resolution Image Synthesis With Latent Diffusion Models},
  author={Rombach, Robin and Blattmann, Andreas and Lorenz, Dominik and Esser, Patrick and Ommer, Björn},
  booktitle={Proceedings of the IEEE/CVF Conference on Computer Vision and Pattern Recognition},
  pages={10684--10695},
  year={2022}
}

@article{naik2023social,
  title={Social Biases Through the Text-to-Image Generation Lens},
  author={Naik, Ranjita and Nushi, Besmira},
  journal={arXiv Preprint arXiv:2304.06034},
  year={2023}
}

@article{DBLP:journals/corr/abs-2112-10752,
  author       = {Rombach, Robin and Blattmann, Andreas and Lorenz, Dominik and Esser, Patrick and Ommer, Björn},
  title        = {High-Resolution Image Synthesis With Latent Diffusion Models},
  journal      = {CoRR},
  volume       = {abs/2112.10752},
  year         = {2021},
  url          = {https://arxiv.org/abs/2112.10752},
  eprinttype   = {arXiv},
  eprint       = {2112.10752},
  bibsource    = {dblp computer science bibliography, https://dblp.org}
}

@misc{dalle3,
    author= {OpenAI},
    title = {DALL·E 3 System Card},
    note  = {\url{https://cdn.openai.com/papers/DALL_E_3_System_Card.pdf}},
    year={2023},
    Accessed={13-03-2025}
}

@misc{dalle33,
    author= {OpenAI},
    title = {DALL·E 3},
    note  = {\url{https://openai.com/dall-e-3}},
    year={2023},
    Accessed={13-03-2025}
}

@article{podell2023sdxl,
  title={SDXL: Improving Latent Diffusion Models for High-Resolution Image Synthesis},
  author={Podell, Dustin and English, Zion and Lacey, Kyle and Blattmann, Andreas and Dockhorn, Tim and Müller, Jonas and Penna, Joe and Rombach, Robin},
  journal={arXiv Preprint arXiv:2307.01952},
  year={2023}
}

@article{caliskan2017semantics,
  title={Semantics Derived Automatically From Language Corpora Contain Human-Like Biases},
  author={Caliskan, Aylin and Bryson, Joanna J and Narayanan, Arvind},
  journal={Science},
  volume={356},
  number={6334},
  pages={183--186},
  year={2017},
  publisher={American Association for the Advancement of Science}
}

@article{greenwald1998measuring,
  title={Measuring Individual Differences in Implicit Cognition: The Implicit Association Test},
  author={Greenwald, Anthony G and McGhee, Debbie E and Schwartz, Jordan LK},
  journal={Journal of Personality and Social Psychology},
  volume={74},
  number={6},
  pages={1464},
  year={1998},
  publisher={American Psychological Association}
}

@inproceedings{orgad2023editing,
  title={Editing Implicit Assumptions in Text-to-Image Diffusion Models},
  author={Orgad, Hadas and Kawar, Bahjat and Belinkov, Yonatan},
  booktitle={Proceedings of the IEEE/CVF International Conference on Computer Vision},
  pages={7053--7061},
  year={2023}
}

@article{wan2024survey,
  title={Survey of Bias in Text-to-Image Generation: Definition, Evaluation, and Mitigation},
  author={Wan, Yixin and Subramonian, Arjun and Ovalle, Anaelia and Lin, Zongyu and Suvarna, Ashima and Chance, Christina and Bansal, Hritik and Pattichis, Rebecca and Chang, Kai-Wei},
  journal={arXiv Preprint arXiv:2404.01030},
  year={2024}
}

@inproceedings{prerak2024addressing,
  title={Addressing Bias in Text-to-Image Generation: A Review of Mitigation Methods},
  author={Prerak, Shah},
  booktitle={2024 Third International Conference on Smart Technologies and Systems for Next Generation Computing (ICSTSN)},
  pages={1--6},
  year={2024},
  organization={IEEE}
}

@misc{bpdefinition,
    author= {American Psychiatric Association},
    title = {What Are Bipolar Disorders?},
    note  = {\url{https://www.psychiatry.org/patients-families/bipolar-disorders/what-are-bipolar-disorders}},
    year={2025},
    Accessed={07-08-2025}
}

@misc{addefinition,
    author= {American Psychiatric Association},
    title = {What are Anxiety Disorders?},
    note  = {\url{https://www.psychiatry.org/Patients-Families/Anxiety-Disorders/What-are-Anxiety-Disorders}},
    year={2025},
    Accessed={07-08-2025}
}

@misc{dedefinition,
    author= {American Psychiatric Association},
    title = {What are Depression?},
    note  = {\url{https://www.psychiatry.org/Patients-Families/Depression/What-Is-Depression}},
    year={2025},
    Accessed={07-08-2025}
}

@misc{ICD-11,
    author= {World Health Organization},
    title = {Bipolar or Related Disorders},
    note  = {\url{https://icd.who.int/browse/2025-01/mms/en#613065957}},
    year={2025},
    Accessed={13-03-2025}
}

@misc{ICD-11-de,
    author= {World Health Organization},
    title = {Depressive Disorders},
    note  = {\url{https://icd.who.int/browse/2025-01/mms/en#1563440232}},
    year={2025},
    Accessed={08-08-2025}
}

@misc{ICD-11-ad,
    author= {World Health Organization},
    title = {Anxiety or Fear-Related Disorders},
    note  = {\url{https://icd.who.int/browse/2025-01/mms/en#1336943699}},
    year={2025},
    Accessed={08-08-2025}
}

@book{american2013diagnostic,
  title={Diagnostic and Statistical Manual of Mental Disorders: DSM-5},
  author={American Psychiatric Association and others},
  year={2013},
  publisher={American Psychiatric Association}
}

@misc{gpt-4o-image,
    author= {OpenAI},
    title = {Introducing 4o Image Generation},
    note  = {\url{https://openai.com/index/introducing-4o-image-generation/}},
    year={2025},
    Accessed={07-08-2025}
}

@article{fan2025lexam,
  title={LEXam: Benchmarking Legal Reasoning on 340 Law Exams},
  author={Fan, Yu and Ni, Jingwei and Merane, Jakob and Tian, Yang and Hermstr{\"u}wer, Yoan and Huang, Yinya and Akhtar, Mubashara and Salimbeni, Etienne and Geering, Florian and Dreyer, Oliver and Brunner, Daniel and Leippold, Markus and Sachan, Mrinmaya and Stremitzer, Alexander and Engel, Christoph and Ash, Elliott and Niklaus, Joel},
  journal={arXiv preprint arXiv:2505.12864},
  year={2025}
}

@inproceedings{wolfe-caliskan-2022-contrastive,
    title = "Contrastive Visual Semantic Pretraining Magnifies the Semantics of Natural Language Representations",
    author = "Wolfe, Robert  and
      Caliskan, Aylin",
    editor = "Muresan, Smaranda  and
      Nakov, Preslav  and
      Villavicencio, Aline",
    booktitle = "Proceedings of the 60th Annual Meeting of the Association for Computational Linguistics (Volume 1: Long Papers)",
    month = may,
    year = "2022",
    address = "Dublin, Ireland",
    publisher = "Association for Computational Linguistics",
    url = "https://aclanthology.org/2022.acl-long.217/",
    doi = "10.18653/v1/2022.acl-long.217",
    pages = "3050--3061"
}

@inproceedings{camacho-collados-etal-2022-tweetnlp,
    title = "{T}weet{NLP}: Cutting-Edge Natural Language Processing for Social Media",
    author = "Camacho-collados, Jose  and
      Rezaee, Kiamehr  and
      Riahi, Talayeh  and
      Ushio, Asahi  and
      Loureiro, Daniel  and
      Antypas, Dimosthenis  and
      Boisson, Joanne  and
      Espinosa Anke, Luis  and
      Liu, Fangyu  and
      Mart{\'i}nez C{\'a}mara, Eugenio",
    editor = "Che, Wanxiang  and
      Shutova, Ekaterina",
    booktitle = "Proceedings of the 2022 Conference on Empirical Methods in Natural Language Processing: System Demonstrations",
    month = dec,
    year = "2022",
    address = "Abu Dhabi, UAE",
    publisher = "Association for Computational Linguistics",
    url = "https://aclanthology.org/2022.emnlp-demos.5/",
    doi = "10.18653/v1/2022.emnlp-demos.5",
    pages = "38--49"
}

@inproceedings{loureiro-etal-2022-timelms,
    title = "{T}ime{LM}s: Diachronic Language Models from {T}witter",
    author = "Loureiro, Daniel  and
      Barbieri, Francesco  and
      Neves, Leonardo  and
      Espinosa Anke, Luis  and
      Camacho-collados, Jose",
    booktitle = "Proceedings of the 60th Annual Meeting of the Association for Computational Linguistics: System Demonstrations",
    month = may,
    year = "2022",
    address = "Dublin, Ireland",
    publisher = "Association for Computational Linguistics",
    url = "https://aclanthology.org/2022.acl-demo.25",
    doi = "10.18653/v1/2022.acl-demo.25",
    pages = "251--260"
}

@inproceedings{fan2025medium,
  title={The Medium Is Not the Message: Deconfounding Document Embeddings via Linear Concept Erasure},
  author={Fan, Yu and Tian, Yang and Ravfogel, Shauli and Sachan, Mrinmaya and Ash, Elliott and Hoyle, Alexander Miserlis},
  booktitle={Proceedings of the 2025 Conference on Empirical Methods in Natural Language Processing},
  pages={32112--32131},
  year={2025}
}

@misc{prolific,
  author       = {{Prolific}},
  title        = {Prolific},
  howpublished = {\url{https://www.prolific.com/}},
  note         = {Accessed: 2025-08-27}
}

@inproceedings{urman2025cultural,
  author    = {Aleksandra Urman and Elsa Lichtenegger and Azza Bouleimen and Robin Forsberg and Corinna Hertweck and Desheng Hu and Stefania Ionescu and Kshitijaa Jaglan and Salima Jaoua and Nicol\`o Pagan and Anik\'o Hann\'ak and Joachim Baumann},
  title     = {The World We See Through AI's Eyes: U.S. Cultural Dominance in Text-to-Image Generation},
  booktitle = {Proceedings of Fourth European Workshop on Algorithmic Fairness (EWAF'25)},
  series    = {Proceedings of Machine Learning Research},
  year      = {2025}
}

@inproceedings{10.1145/3757887.3763012,
author = {Bennett, Cynthia L. and Kane, Shaun K. and Harrington, Christina N.},
title = {Toward Community-Led Evaluations of Text-to-Image AI Representations of Disability, Health, and Accessibility},
year = {2025},
isbn = {9798400721403},
publisher = {Association for Computing Machinery},
address = {New York, NY, USA},
url = {https://doi.org/10.1145/3757887.3763012},
doi = {10.1145/3757887.3763012},
booktitle = {Proceedings of the 5th ACM Conference on Equity and Access in Algorithms, Mechanisms, and Optimization},
pages = {256–270},
numpages = {15},
location = {
},
series = {EAAMO '25}
}

\appendix

\section{DALL·E 3 API Parameters}
\label{appendix:dalle3}
We used the OpenAI API with the following parameters:

\begin{verbatim}
model = "dall-e-3"
prompt = "photo of a person with a disability"
n = 1
quality = "standard"
size = "1024x1024"
response_format = "b64_json"
\end{verbatim}

\end{document}